\documentclass[10pt,twocolumn,letterpaper]{article}

\usepackage{cvpr}
\usepackage{times}
\usepackage{epsfig}
\usepackage{graphicx}
\usepackage{amsmath}
\usepackage{amssymb}
\usepackage{algorithm}
\usepackage{algorithmic}
\usepackage{array}
\usepackage{booktabs}
\usepackage[margin=1in]{geometry}
\usepackage{tikz}
\usepackage{xcolor}

\usetikzlibrary{shapes.geometric, arrows, positioning, fit, shadows, decorations.pathreplacing}
\tikzset{
  arrow/.style = {->, >=stealth, thick, draw=accentcolor},
}

\definecolor{lightblue}{RGB}{173, 216, 230}
\definecolor{lightgreen}{RGB}{144, 238, 144}
\definecolor{lightyellow}{RGB}{255, 255, 224}
\definecolor{lightgray}{RGB}{240, 240, 240}
\definecolor{darkgray}{RGB}{64, 64, 64}
\definecolor{datacolor}{RGB}{230, 240, 250}
\definecolor{processcolor}{RGB}{255, 248, 220}
\definecolor{outputcolor}{RGB}{240, 255, 240}
\definecolor{accentcolor}{RGB}{70, 130, 180}

\newcommand{\websight}{\textsc{WebSight}}

\usepackage[breaklinks=true,bookmarks=false]{hyperref}

\cvprfinalcopy 

\def\cvprPaperID{****} 

\begin{document}

\title{\websight: A Vision-First Architecture for Robust Web Agents}

\author{Tanvir Bhathal*\\
Stanford University\\
{\tt\small tanvirb@stanford.edu}
\and
Asanshay Gupta*\\
Stanford University\\
{\tt\small asanshay@stanford.edu}
}

\maketitle

\begin{abstract}
    We introduce \websight, a vision-based autonomous web agent, designed to interact with web environments purely through visual perception, eliminating dependence on HTML or DOM-based inputs. Central to our approach we introduce our new model, \websight-7B, a fine-tuned vision-language model optimized for UI element interaction, trained using LoRA on a web-focused subset of the Wave-UI-25K dataset. \websight\ integrates this model into a modular multi-agent architecture, comprising planning, reasoning, vision-action, and verification agents, coordinated through an episodic memory mechanism. 

    \websight-7B achieves a top-1 accuracy of {58.84\%} on the Showdown Clicks benchmark, outperforming several larger generalist models while maintaining lower latency. The full \websight\ agent achieves a {68.0\%} success rate on the WebVoyager benchmark, surpassing systems from labs such as OpenAI (61.0\%) and HCompany (Runner H, 67.0\%). Among tasks completed, \websight\ answers correctly {97.14\%} of the time, indicating high precision. Together, \websight\ and \websight-7B establish a new standard for interpretable, robust, and efficient visual web navigation.
\end{abstract}

\section{Introduction}

Autonomous web agents capable of performing complex web navigation tasks—such as automated form filling, online shopping, and dynamic information retrieval—have emerged as a critical area of study within artificial intelligence. Over recent years, substantial progress has been driven by leveraging large language models (LLMs), enabling browser agents to interpret web content primarily through textual representations like HTML code, Document Object Model (DOM) trees, and accessibility metadata~\cite{nakano2021webgpt, xu2023wizardlm, yang2023sphynx}. Despite their successes, these approaches present critical challenges when confronted with real-world scenarios. Specifically, websites frequently feature incomplete or incorrect metadata, dynamic layouts, and complex designs that degrade the reliability of structurally-dependent agents~\cite{sun2022recovering, xu2022focal}. Furthermore, the computational demands of processing extensive textual and structured inputs limit their scalability and interpretability, hindering practical deployment~\cite{zhang2023autoagents, li2024agentbench, ganguli2022predictability}.

\begin{figure}[t]
    \centering
    \includegraphics[width=1\linewidth]{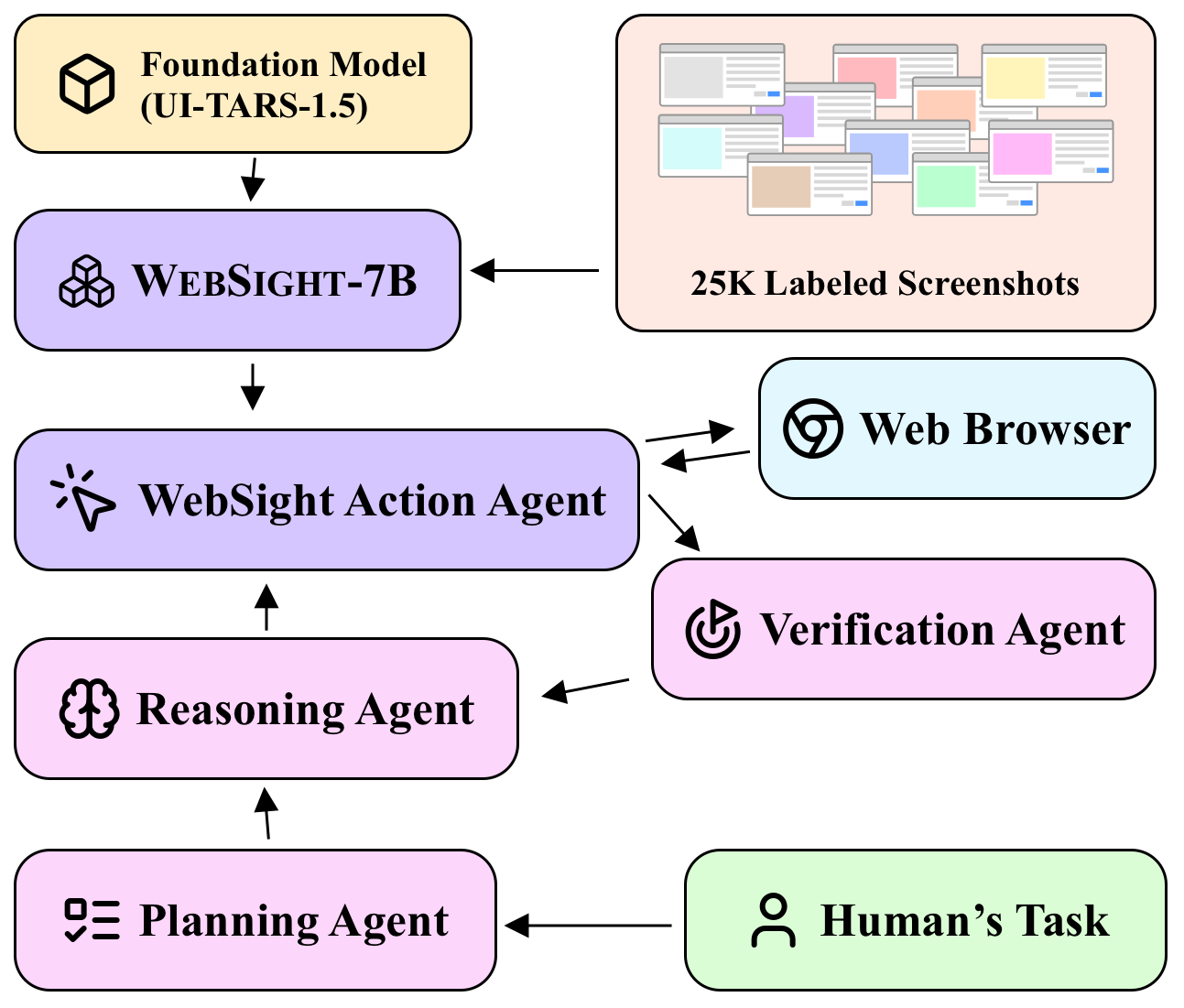}
    \caption{The \websight\  Architecture}
    \label{fig:websight-loop}
\end{figure}

In contrast, human users rely almost exclusively on visual perception, effortlessly recognizing actionable interface elements like buttons, input fields, and navigation bars based solely on visual layout and design cues, irrespective of underlying metadata~\cite{donker2002visual, zhang2021screen2words}. Inspired by this natural modality, we present two tightly integrated contributions: \websight, a vision-based autonomous browser agent; and \websight-7B, a task-specific vision-language model trained to identify and interact with user interface elements directly from rendered web screenshots.

The \websight\ agent is built on a robust multi-agent orchestration framework that mimics human cognitive processes.  At its core are dedicated planning agents, which devise comprehensive strategies outlining task sequences and continually track progression towards goals~\cite{kaelbling1998planning, russell2010artificial}. Concurrently, specialized reasoning agents perform detailed analyses to pinpoint and articulate precise subsequent actions necessary for task advancement~\cite{bommasani2021opportunities, ganguli2022predictability}. The critical component of our architecture, the vision agent, leverages the \websight-7B model~\cite{qin2025ui, hu2022lora} to make decisions on actions in the browser. Post-action, dedicated verification agents rigorously evaluate the resultant webpage state changes to ascertain accuracy and effectiveness of each interaction~\cite{sun2022recovering, xu2022focal}. This integrative process is further bolstered by an episodic memory mechanism, dynamically updating and iteratively refining agent strategies through continuous verification cycles until completion criteria are achieved~\cite{shi2022learning, li2024agentbench, zhang2023autoagents}. Collectively, this orchestrated multi-agent architecture closely mirrors human cognitive and perceptual workflows, significantly enhancing interpretability, adaptability, and robustness~\cite{donker2002visual, zhang2021screen2words}.


At the core of this system, \websight-7B is a fine-tuned version of UI-TARS that we adapt to the web domain \cite{qin2025ui}. Trained on augmented web-based GUI datasets using LoRA fine-tuning, \websight-7B demonstrates substantial improvements in English-language understanding and precision UI interaction over generalist vision-language models \cite{hu2022lora}. We show that the combination of architectural modularity and a domain-specialized foundation model enables both strong performance and practical deployability.

To validate our contributions, we evaluate \websight\ and \websight-7B on two challenging benchmarks: WebVoyager and Showdown Clicks~\cite{he2024webvoyager, skyvern2024webvoyager, showdown2025}. \websight\ achieves a 68.0\% success rate on WebVoyager, outperforming comparable agents by 1–7 percentage points. On the Showdown Clicks benchmark, \websight-7B attains a top-1 accuracy of 58.84\%, surpassing a range of larger general-purpose vision-language models by 4–7 points. These results underscore the benefits of a vision-first architecture and domain-specialized model design for effective and efficient web interaction.



Through \websight, we introduce a dual innovation: a vision-first web agent and an accompanying domain-optimized vision-language model. Together, advance autonomous web agent research by explicitly integrating human-like visual perception into agent design. Our approach offers a path toward robust, interpretable, and computationally efficient autonomous agents capable of generalizing across diverse web environments, aligning closely with human browsing behavior and intuitions.

We publicly share \websight's code on \href{https://github.com/SuperAce100/websight}{Github}: \url{https://github.com/SuperAce100/websight} and \websight-7B on HuggingFace: \url{https://huggingface.co/tanvirb/websight-7B}.


\section{Related Works}

Early web navigation agents predominantly relied on reinforcement learning techniques, demonstrating basic task execution capabilities~\cite{lauer2004learning, tesauro2005online, branavan2009reinforcement}. The advent of deep reinforcement learning further enhanced agent capabilities, enabling more sophisticated interaction strategies~\cite{mnih2015human, schulman2017proximal, hessel2018rainbow, vinyals2019grandmaster}. However, these agents often required extensive training environments and struggled with generalization to unseen websites.

More recent advancements in large language models (LLMs) such as GPT series~\cite{brown2020language, ouyang2022training, openai2023gpt4} have significantly improved autonomous agents' abilities to generalize and handle complex instructions~\cite{zhou2023webarena, liu2023webagent, xu2023wizardlm}. Nonetheless, these approaches remain heavily reliant on structured textual inputs such as DOM trees and accessibility metadata, limiting their robustness against incomplete or inaccurate metadata~\cite{sun2022recovering, xu2022focal, zhang2023autoagents, li2024agentbench}.

Parallel to textual methods, computer vision approaches have significantly advanced, particularly in semantic segmentation and object detection~\cite{he2017mask, redmon2018yolov3, chen2017deeplab, carion2020endtoend, zhang2020resnest}. Recent progress in visual transformer architectures and self-supervised learning has further pushed the boundary of vision-based perception capabilities~\cite{dosovitskiy2020image, caron2021emerging, radford2021learning, bao2022beit, wang2022image}.

Integrating visual perception into web navigation has also garnered attention. Early work utilized vision heuristics for webpage understanding~\cite{cai2001extracting, kumar2007extracting, dong2010towards}. More recent methods have employed deep learning models to directly interpret visual webpage layouts, significantly improving navigation robustness~\cite{zhang2021screen2words, kirillov2023segment, ravi2024sam2, liu2023webui}.

Our multi-agent orchestration approach builds upon foundational theories in multi-agent systems and planning~\cite{kaelbling1998planning, russell2010artificial, wooldridge2009introduction}, cognitive architectures~\cite{anderson2009can, laird2012soar}, episodic memory integration~\cite{blundell2016model, pritzel2017neural}, and human-inspired verification strategies~\cite{mnih2015human, silver2016mastering, lecun2022path} to achieve more robust, interpretable, and efficient autonomous web agents.

\section{Data}

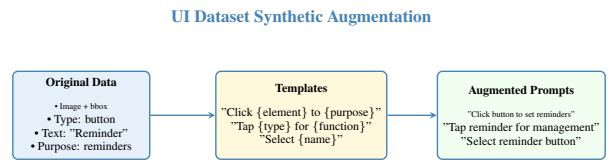
\begin{figure}[ht]
\centering
\scalebox{0.65}{
\begin{tikzpicture}[
    node distance=1.8cm,
    every node/.style={align=center, font=\scriptsize},
    mainbox/.style={rectangle, draw=accentcolor, thick, rounded corners=3pt, minimum height=1.8cm}
]

\node[font=\bfseries, color=accentcolor] at (0, 4) {UI Dataset Synthetic Augmentation};

\node[mainbox, fill=datacolor, minimum width=2.8cm] (input) at (-4.5, 2) {
    \textbf{Original Data}\\[0.2cm]
    \tiny
    • Image + bbox\\
    • Type: button\\
    • Text: "Reminder"\\
    • Purpose: reminders
};

\node[mainbox, fill=processcolor, minimum width=2.8cm] (process) at (0, 2) {
    \textbf{Templates}\\[0.2cm]
    "Click \{element\} to \{purpose\}"\\
    "Tap \{type\} for \{function\}"\\
    "Select \{name\}"
};

\node[mainbox, fill=outputcolor, minimum width=2.8cm] (output) at (4.5, 2) {
    \textbf{Augmented Prompts}\\[0.2cm]
    \tiny
    "Click button to set reminders"\\
    "Tap reminder for management"\\
    "Select reminder button"
};

\draw[arrow] (input.east) -- (process.west) ;
\draw[arrow] (process.east) -- (output.west);

\end{tikzpicture}
}
\caption{Synthetic augmentation pipeline for UI vision-language model training. Original structured UI data is processed through natural language templates to generate diverse instruction variations.}
\label{fig:ui_augmentation}
\end{figure}

We utilize AgentSea's Wave-UI-25K dataset~\cite{waveui25k}. Wave-UI-25K comprises of 24,978 entries. Of this, we use the subset of 22,994 web-based screenshots with annotated browser interaction elements, as we are not focusing on alternative subsets like mobile and desktop. 

Each dataset entry includes a screenshot, the bounding box of a particular UI element within it, OCR extracted text from the image, and a natural language description of the purpose of the element and expected result from interacting with the element.

In Figure ~\ref{fig:ui_augmentation} we optimize the responsiveness of \websight\ to a variety of natural language prompts by synthetically augment the dataset with procedurally generated natural language prompts based on the semantic labels.

Fine-tuning allows the UI-TARS model to refine its attention and enhance task-specific efficacy, leveraging existing visual representation capabilities to robustly handle nuanced web interactions~\cite{devlin2019bert, howard2018universal, raffel2020exploring, qin2025ui}. We also ensure the model's proficiency in delivering precise English instructions and interactions specifically tailored for web environments (Section \ref{sec:finetune}).

We evaluate \websight\ using established benchmarks such as Showdown Clicks~\cite{showdown2025}, which provides a robust test to determine accuracy of click locations of a VLM used in a Browser Agent. Additionally, we utilize Skyvern AI's filtered WebVoyager dataset~\cite{skyvern2024webvoyager}, an enhanced evaluation set derived from the original WebVoyager dataset~\cite{he2024webvoyagerbuildingendtoendweb}. Skyvern's dataset specifically removes tasks deemed impossible due to outdated web page structures, ensuring more accurate and meaningful performance assessments.


\begin{figure}[h]\
\centering
\includegraphics[width=0.48\textwidth]{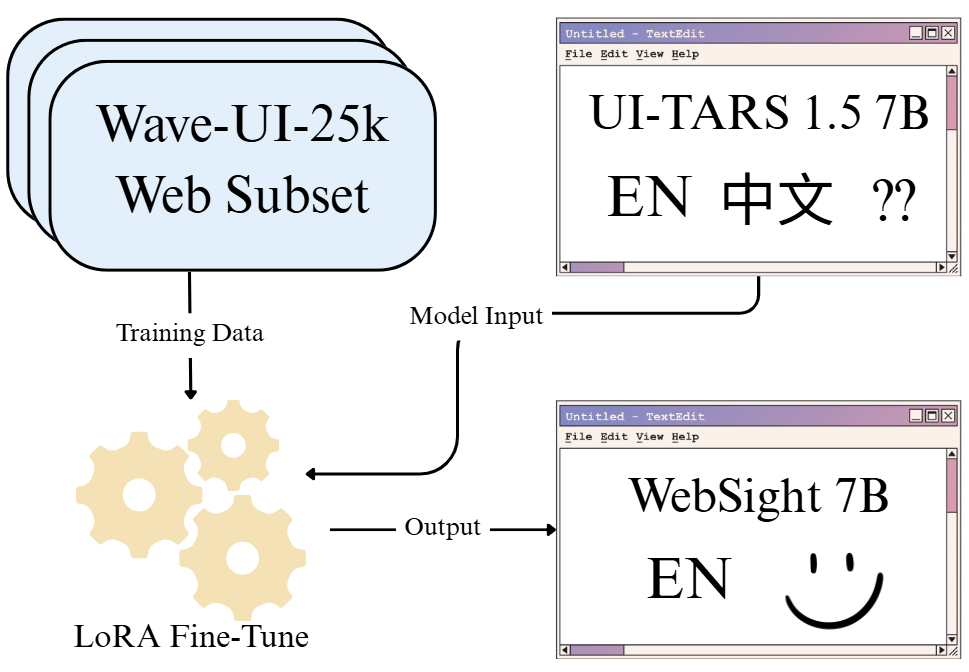}
\caption{UI-TARS LoRA Fine Tuning Process to develop \websight-7B. Leverages Web Subset of Wave-UI-25k to make UI-TARS English and browser specific.}
\label{fig:lora}
\end{figure}

\section{Methods}

\websight\ is a visually-grounded multi-agent framework specifically designed to emulate human browser navigation behaviors. Unlike traditional LLM-based agents that rely heavily on structured textual inputs such as HTML or accessibility metadata, \websight\ leverages purely visual inputs—webpage screenshots—to interact effectively with web environments.

\subsection{Preliminaries}

\textbf{Baselines} \; \; To effectively benchmark \websight's performance we determine the effectiveness of both \websight-7B and the \websight\ agent. For \websight-7B we compare its performance against state-of-the-art (SOTA) Vision-Language Models (VLMs), all of which are transformer-based and natively support multimodal input. These include models such as GPT-4o, Gemini 2.0 Flash, and Claude 3.7 Sonnet, as well as other large-scale VLMs and vision-capable LLMs. For the \websight\ agent, we compare it to other SOTA web agents, whether they utilize vision or not. These agents include, Browserable, Browser Use, Skyvern 2.0, Agent-E, as well as many more by other cutting edge startups. 

\textbf{Problem Definition} \; \; We formalize the problem using the following notation. Given an initial task $\beta$ specified by the user, we assign it to the agent’s episodic memory $\mathcal{M}$ and initialize the task state as $\mathcal{T}_0=\beta$. Based on $\mathcal{T}_0$, the system generates a plan $\mathcal{P}$ that \websight\ executes. To select actions $a_t$, the reasoning agent $f$ processes the current status inputs, incorporating relevant information from $\mathcal{V}$. Once $f$ determines that the task is complete, we mark the final state as $\mathcal{T}_{complete}$.

\subsection{\textbf{\websight}-7B}
We present \websight-7B, a Vision Language Model trained for targeted interaction with UI elements on screen given natural language prompts. We integrated this model into our subsequent multi-agent pipeline. We detail our process of developing the model below:

\subsubsection{Fine-tuning}\label{sec:finetune}
In Figure~\ref{fig:lora}, we fine-tune UI-TARS-1.5-7B~\cite{qin2025ui} using the Wave-UI-25K dataset~\cite{waveui25k}. Each training sample consists of a procedurally generated natural language instruction and a website screenshot. Originally, UI-TARS outputs frequently in Chinese and is too generalized to GUI interactions, leading to performance issues in English settings and the web domain. Our fine-tuning explicitly enhances its capability to respond to focused English instructions and accurately perform browser-specific interactions. Fine-tuning improves domain-specific performance by adapting general-purpose visual representations to specialized tasks~\cite{howard2018universal}.

We fine-tune UI-TARS using LLaMA-Factory ~\cite{zheng2024llamafactory} with bf16 quantization and a LoRA approach on the attention layers for faster training. Our fine-tuning leverages supervised learning with a cross-entropy loss for instruction generation. Within this setting, training on 2 NVIDIA L40S GPUs takes about 6 hours.

Further, from Figure~\ref{fig:tloss} we see that a majority of the learning is done in the beginning of the process, with the returns tapering quickly. This is due to the web browser and english support already existing in UI-TARS and the fine tuning process only supporting and enhancing this support. Further, the process adds a limited amount of important new tokens. We detail this in more detail in the appendix Section 7.2.


\begin{figure}[h]
    \centering
    \includegraphics[width=0.5\textwidth]{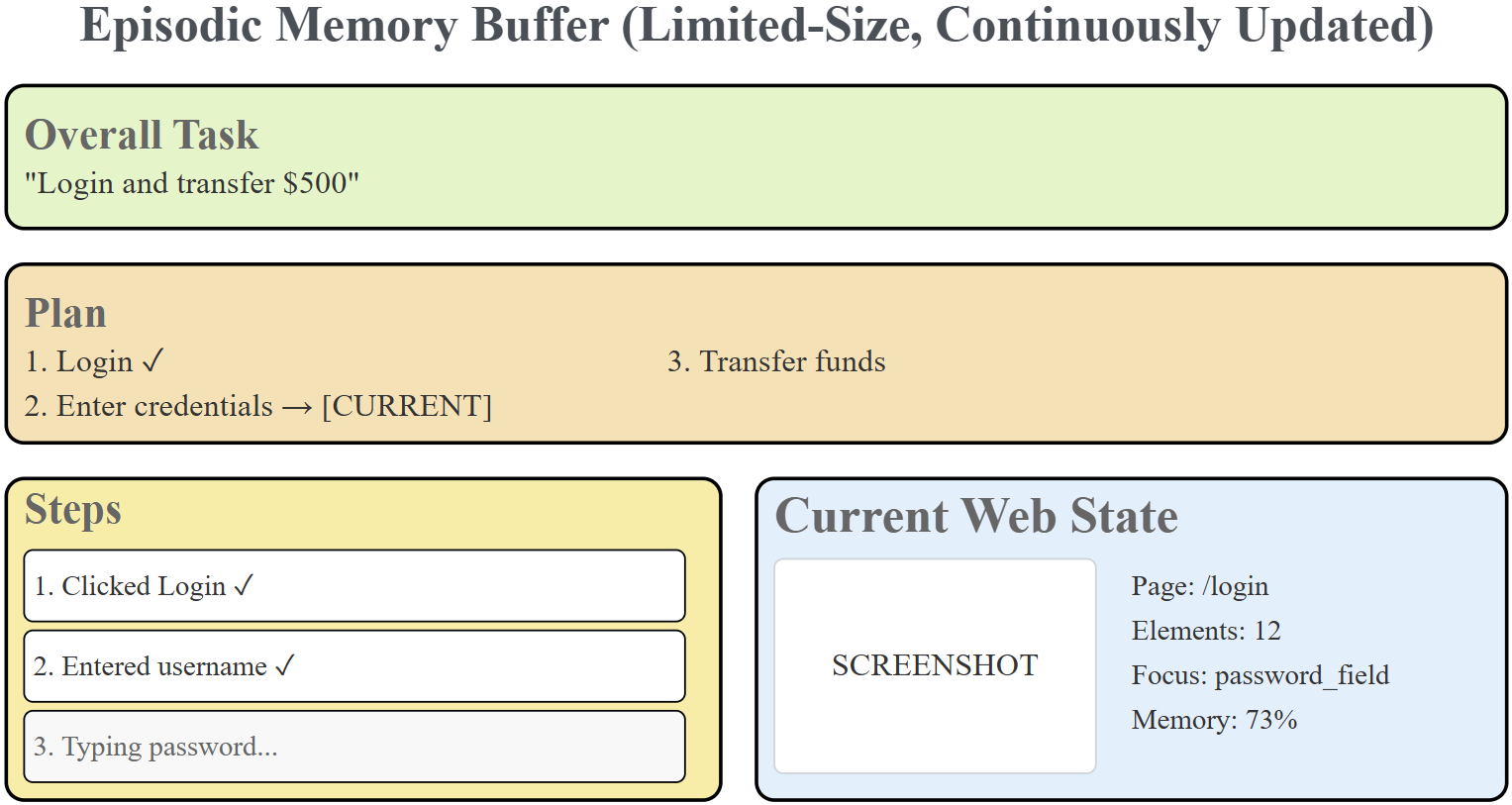}
    \caption{Episodic Memory Diagram dictates Task, Planning, Step Tracking, and Current Web State that form the Episodic Memory.}
    \label{fig:episodic}
\end{figure}

\subsection{Multi-Agent Framework}
\websight\ implements a modified version the ReAct agent framework augmented with a planning agent and separated reasoning and action sub-steps visualized in Algorithm ~\ref{algo1} ~\cite{yao2023reactsynergizingreasoningacting}:

\begin{enumerate}
    \item\textbf{Planning Agents.} These agents are responsible for constructing high-level action plans based on user instructions or task objectives. Leveraging classic chain-of-thought planning methodologies~\cite{wei2023chain}, our planning agents formulate task sequences that provide long-term context for the reasoning agents. These plans are dynamically updated based on feedback from verification agents.

    \item\textbf{Reasoning Agents.} Situated below planning agents, reasoning agents determine precise next-step interactions necessary to progress the overall task. Utilizing transformer-based reasoning architectures~\cite{devlin2019bert, raffel2020exploring}, these agents translate high-level plans into specific actionable steps such as "click the login button" or "fill in a text box."

    \item\textbf{Action Agent.} Central to \websight\ is our custom trained \websight-7B based agent, which interprets semantic instructions from the reasoning agents and translates them into visual interactions directly on screenshots. 

    \item\textbf{Verification Agents.} After the vision agent executes an action, verification agents rigorously evaluate the resulting changes in the webpage state. Utilizing visual reasoning, these agents verify task progress, update memory, and determine what the next step to be taken is--which was seen.
\end{enumerate}

\subsection{Episodic Memory} Integral to human-like interaction, in Figure~\ref{fig:episodic} we incorporate a short-term episodic memory that records recent interactions and webpage states \cite{sumers2024cognitivearchitectureslanguageagents}. This memory allows our agents to iteratively refine action strategies, detect errors quickly, and prevent repetitive mistakes~\cite{blundell2016model, pritzel2017neural}. The episodic memory is structured as a limited-size buffer storing recent interaction-action-outcome tuples, continuously updated and pruned based on task relevance.

\begin{algorithm}
\caption{\websight\ Agent Loop}
\begin{algorithmic}[1]
\label{algo1}
\STATE Initialize episodic memory $\mathcal{M} = \emptyset$ and  task state $\mathcal{T}_0$\STATE \textbf{Planning:} Create plan $\mathcal{P} = \{p_1, p_2, \ldots, p_n\}$
\WHILE{$\mathcal{T}_t \neq \mathcal{T}_{complete}$}
    \STATE \textbf{Reasoning:} Determine next action \\$a_t = f(\mathcal{P}, \mathcal{M}, \mathcal{T}_t)$
    \STATE \textbf{Action:} Execute \websight\ visual action $\mathcal{V}(a_t)$\STATE \textbf{Verification:} Assess visual changes $\Delta\mathcal{V}_t$ and compute $\mathcal{T}_{t+1}$
    \STATE \textbf{Episodic Memory Update:}\\ $\mathcal{M} = \mathcal{M} \cup \{(a_t, \Delta\mathcal{V}_t, \mathcal{T}_t, \mathcal{T}_{t+1})\}$
    \IF{$\mathcal{T}_{t+1} \succ \mathcal{T}_t$}
        \STATE $\mathcal{T}_t \leftarrow \mathcal{T}_{t+1}$
    \ELSE
        \STATE $\mathcal{P} \leftarrow \text{UpdatePlan}(\mathcal{P}, \mathcal{M}, \mathcal{T}_t)$
    \ENDIF
\ENDWHILE
\RETURN $\mathcal{T}_{complete}$
\end{algorithmic}
\end{algorithm}

\subsection{Alternative Approaches Considered}
We initially explored purely monolithic architectures, such as directly using large multimodal transformer models~\cite{radford2021learning, dosovitskiy2020image}, to handle both perception and reasoning simultaneously. However, we found that decomposing perception and reasoning into specialized agents provided greater interpretability, modularity, and accuracy. Additionally, preliminary experiments indicated that fine-tuning domain-specific vision models resulted in better visual interaction accuracy than generalized multimodal models.

Our multi-agent orchestration framework, combined with targeted fine-tuning strategies, provides a robust and efficient solution, explicitly mirroring human-like browsing behaviors and ensuring practical applicability in diverse web navigation tasks.

\section{Experiments}

We conduct two experiments to evaluate \websight-7B and the \websight\ agent. First, the Showdown-Clicks benchmark, which isolates low-level click accuracy, is used to evaluate \websight-7B compared to other leading VLMs. The WebVoyager benchmark, which measures end-to-end task completion in a realistic multi-page web environment, is used to evaluate the \websight\ browser agent.

\subsection{Showdown/Clicks}

Showdown is a recent suite of offline and online tests for computer-use agents, released by General Agents in early 2025.  The showdown-clicks track contains 5\,679 human-collected left-click events on macOS, with a public dev subset of 557 examples. The benchmark is composed of a wide variety of challenging UI tasks requiring complex skills like understanding icons, processing semantic intent, and acting in situations where there are multiple viable tasks\cite{showdown2025}.

Published baselines vary widely. The SOTA, OpenAI's o3-based CUA reaches 64.27\% top-1 accuracy, whereas a vanilla GPT-4o agent achieves only 5.2\%, underscoring the difficulty of ambiguous UI contexts. Using the same test conditions, \websight-7B\ attains \textbf{58.84\%} accuracy, achieving higher accuracy than VLMs with almost 10x more parameters. Furthermore, \websight-7B served on a single Nvidia H100 GPU is significantly faster than OpenAI's CUA for a slight drop in accuracy.

\begin{table}[ht]
\centering
\scalebox{0.8}{
\begin{tabular}{lcc}
\toprule
\textbf{Model} & \textbf{Top-1 Accuracy (\%)} & \textbf{Latency (ms)} \\
\midrule
\textbf{\websight-7B} & \textbf{58.84} & 2841 \\
OpenAI CUA (o3-based) & 64.27 & 6385 \\
Molmo-72B-0924 & 54.76 & 6599 \\
Claude 3.7 Sonnet & 53.68 & 9656 \\
UI-TARS-72B-SFT & 54.40 & 1977 \\
OmniParser V2 + GPT-4o & 51.71 & 12642 \\
Gemini 2.0 Flash & 33.39 & 3069 \\
Qwen2.5-VL-72B-Instruct & 24.78 & 3790 \\
GPT-4o & 5.21 & 2500 \\
\bottomrule
\end{tabular}
}
\caption{Top-1 Accuracy on the Showdown/Clicks Benchmark \cite{showdown2025}}
\label{tab:showdown_clicks}
\end{table}

\subsection{WebVoyager}

The WebVoyager benchmark is a large-scale, real-world evaluation suite designed to measure the capabilities of autonomous web agents in handling interactive tasks across dynamic websites \cite{he2024webvoyager}. It spans hundreds of user intents on popular domains and tests abilities such as DOM reasoning, form filling, and multi-step navigation. Unlike synthetic tasks, WebVoyager emphasizes natural interaction and robustness in the open web, making it a strong benchmark for evaluating practical utility and generalization in vision-language agents.

Due to computational constraints and to ensure consistent evaluation, we use the filtered subset of 50 tasks introduced by Skyvern \cite{skyvern2024webvoyager}, which excludes outdated live tasks that are no longer possible. Table~\ref{tab:webvoyager-results} shows performance across a variety of state-of-the-art agents evaluated on either the full or filtered WebVoyager set \cite{skyvern2024stateofart}. \websight\ achieves a Success Rate of 68\% on Skyvern's filtered WebVoyager Benchmark \cite{skyvern2024webvoyager}.

\begin{table}[ht]
\centering
\begin{tabular}{lc}
\toprule
\textbf{Agent} & \textbf{Success Rate (\%)} \\
\midrule
\textbf{\websight}           & \textbf{68.0} \\
Browserable                  & 90.4 \\
Browser Use                  & 89.1 \\
Skyvern 2.0                  & 85.8 \\
Claude Computer Use          & 77.5 \\
Agent-E                      & 73.1 \\
Runner H 0.1                 & 67.0 \\
OpenAI Operator (GPT-4o)              & 61.0 \\
WebVoyager Agent             & 57.1 \\
\bottomrule
\end{tabular}
\caption{WebVoyager Benchmark Success Rates \cite{browserable2024webvoyager, skyvern2024stateofart, browseruse2024sota, emergence2024agente, hcompany2024runnerh, claude2024computeruse, openai2024operator}}
\label{tab:webvoyager-results}
\end{table}

From Table~\ref{tab:webvoyager-results} we see that the SOTA Browserable achieves a success rate of 90.4\%. Frontier labs like OpenAI and Anthropic achieve results of 61\% and 77.5\% respectively, highlighting the difficulty of developing a sophisticated browser agent even for labs with many resources.

\subsection{Discussion \& Analysis}

\subsubsection{\textbf{\websight}-7B}


We analyze screenshots of the failures in \websight-7B and identify 3 key failure modes:

\begin{enumerate}
    \item \textbf{Visual Grounding}: While \websight-7B excels at identifying buttons and elements with text labels. Figure \ref{fig:misclick-icon} shows such a use case, where \websight-7B identifies the correct text but misses the interactivity of the icon.
    
    \begin{figure}[b]
        \centering
        \includegraphics[width=0.75\linewidth]{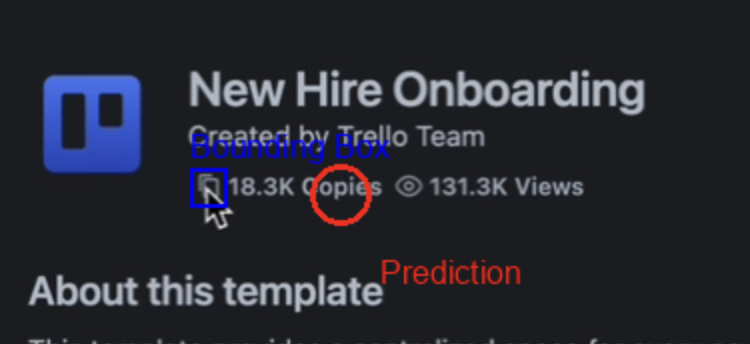}
        \caption{Task: Click on the button to copy}
        \label{fig:misclick-icon}
    \end{figure}
    
    \item \textbf{Extended Action Space}: The action space of \websight-7B extends beyond clicks to include actions like text-input, scrolling, etc. Sometimes, \websight-7B chooses those actions when it does not find a direct answer to the task. Figure \ref{fig:misclick-scroll} shows such a failure mode, where the model chose to scroll to see more posts instea fo clicking the visible one.
    
    \begin{figure}[b]
        \centering
        \includegraphics[width=0.75\linewidth]{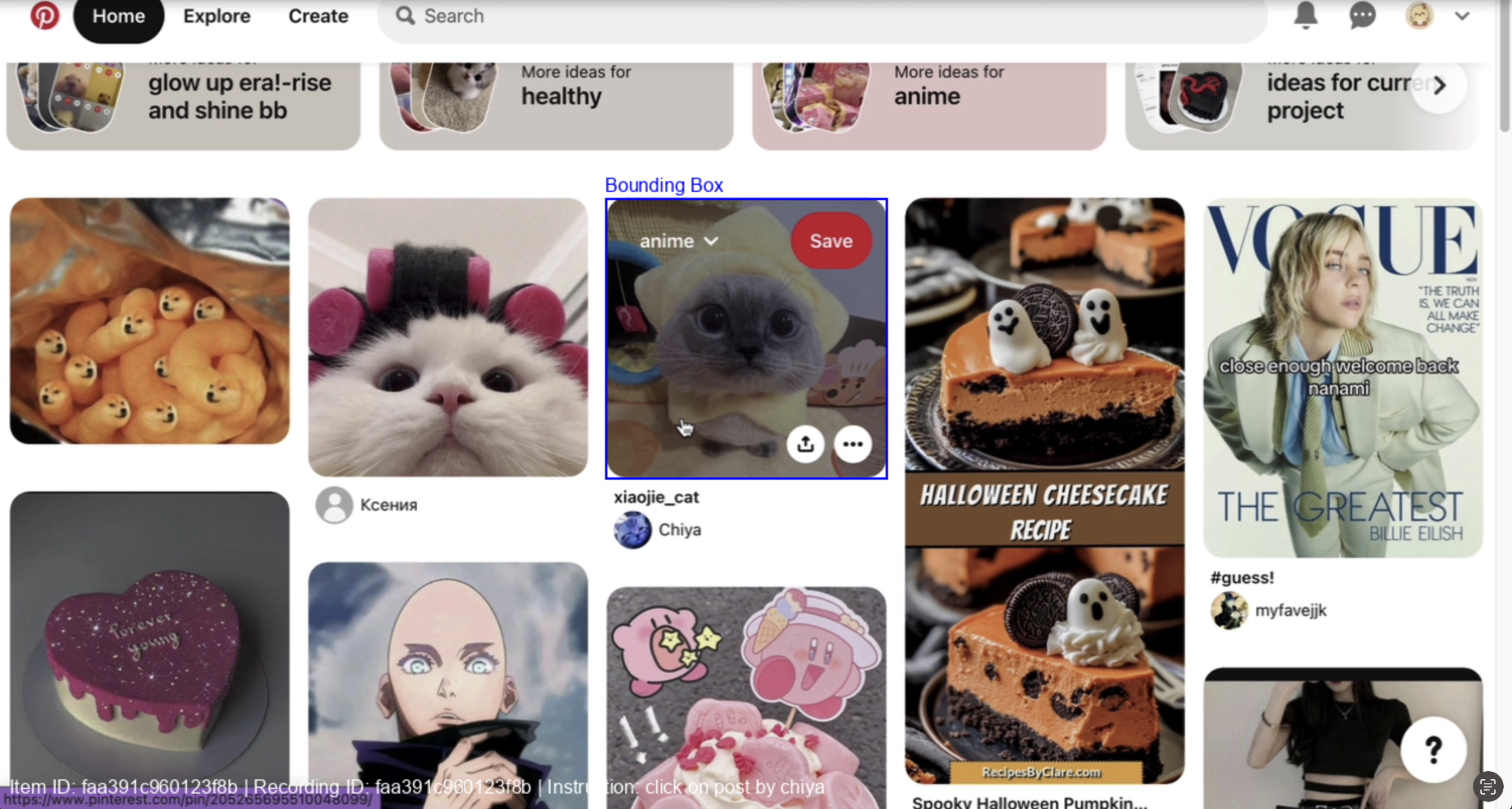}
        \caption{Task: Click on the post by Chiya}
        \label{fig:misclick-scroll}
    \end{figure}
    \item \textbf{Icon Understanding}: \websight-7B also struggles to distinguish between ambiguous icons that may have been out of the training distribution. Figure \ref{fig:misclick-ambig} shows a case where \websight-7B misidentifies an icon and therefore makes an incorrect click.
    
    \begin{figure}[b]
        \centering
        \includegraphics[width=0.75\linewidth]{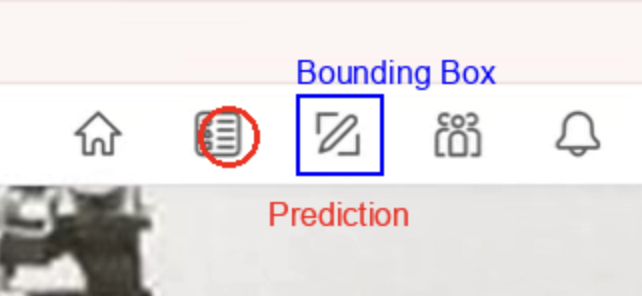}
        \caption{Click on the Answer icon}
        \label{fig:misclick-ambig}
    \end{figure}
    
\end{enumerate}

Two of these failure modes tend to deal with interaction and understanding of icons, which highlights a key weakness in the choice of training data. Appendix 7.3 contains more examples of failures. It is important to note that these failure modes are all fairly uncommon in normal web browsing, and \websight-7B excels at most clicking tasks. 

\subsubsection{\textbf{\websight}\ Agent}

\begin{figure}[h]
    \centering
    \includegraphics[width=0.5\textwidth]{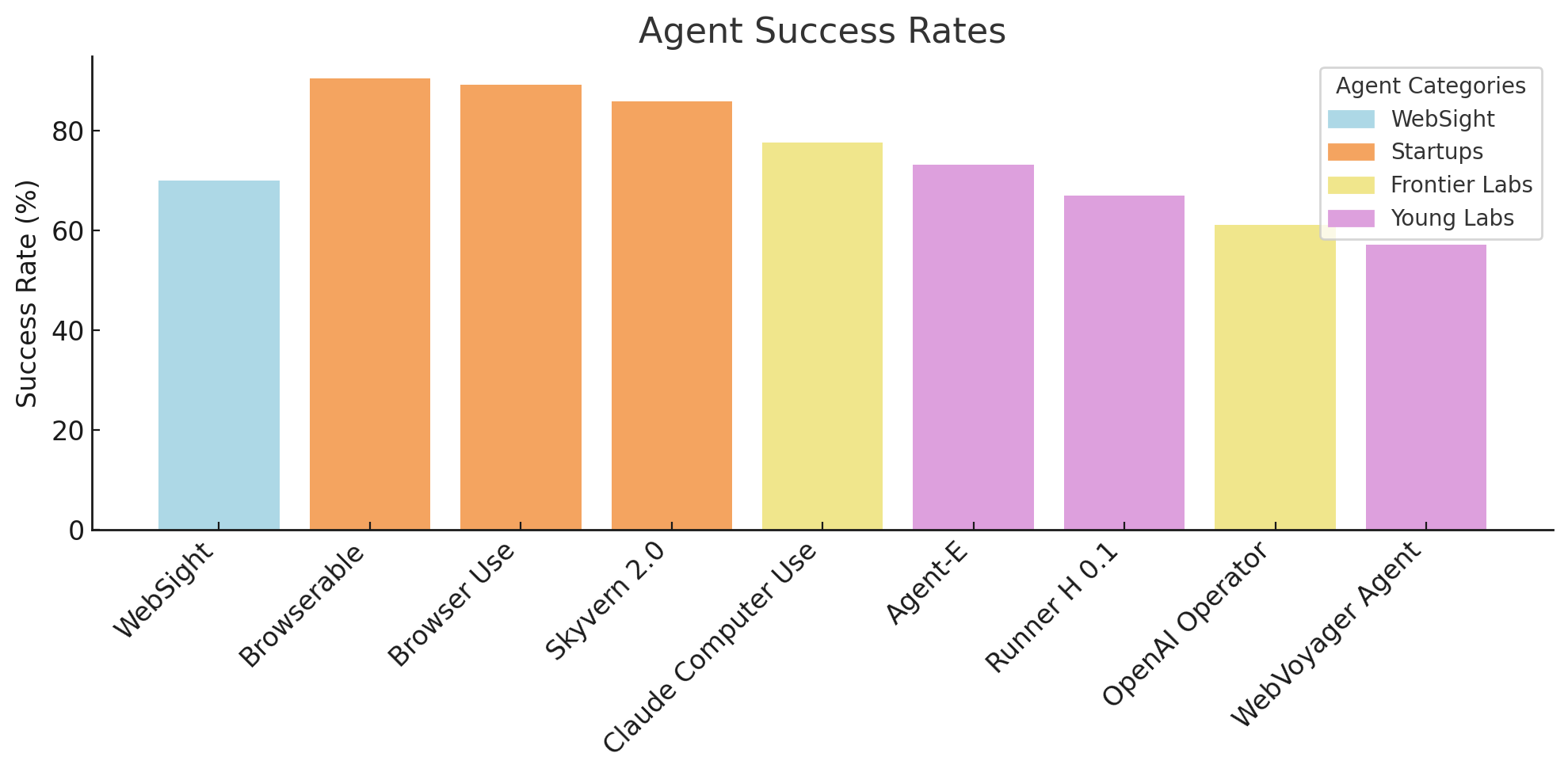}
    \caption{\websight\ performance against other agents}
    \label{fig:agent}
\end{figure}

In Figure~\ref{fig:agent}, we observe that agents developed by startups exhibit the highest success rates. This performance likely reflects a concentrated focus on productization and rapid iteration. In contrast, research-oriented labs tend to pursue broader agendas, which may contribute to comparatively lower performance metrics. Notably, our agent outperforms those from both Frontier Labs and several younger labs, as shown in Table~\ref{tab:webvoyager-results}.

Due to the small size of the Skyvern WebVoyager being only 50 tasks, we manually analyze the 16 tasks that fail to determine insights. We manually test each Planning agent's plan, Reasoning agent's proposed action, and Action agent's step to determine who was at fault, to glean interpretability on where \websight\ could be improved. We know that if at any point \websight\ produced an incorrect result, the Verification Agent has failed, since it did not catch the error.

Upon analysis of trajectories, we see that 15 of the 16 failures were due to the agent timing out, given our 10 minute restriction for \websight\ to complete tasks. We saw that these 15 failures all had infinite loops. Table 3 highlights that the LLMs that power the Planning and Reasoning agent are \websight's weakest links. Some potential ways infinite loops arose were:
\begin{itemize}
    \item Planning agents sending \websight\ into a place with no results, where \websight\ would click a button, navigate backwards and repeat
    \item Reasoning agents repeatedly making the wrong decision, sending \websight\ to a page, navigating backwards, and repeating
    \item Action agents repeatedly being unable to identify buttons or correct click locations as they are too small or fail to understand icons, and repeatedly do nothing
\end{itemize}

\begin{table}[h]
\centering
\label{tab:websight-failure}
\begin{tabular}{l c}
\toprule
\textbf{Component} & \textbf{Failure Count} \\
\midrule
Planning Agent     & 5 \\
Reasoning Agent    & 7 \\
Action Agent       & 3 \\
\bottomrule
\end{tabular}
\caption{Failure Analysis of \websight\ Agent by Component}
\end{table}
 
Further, we understand that of the 35 tasks that \websight\ provided an answer to, 34 were correct. This is a remarkable 97.14\% accuracy, highlighting that \websight\ only provided an answer to the task if it is confident about its answer.

\section{Conclusion}
\subsection{Key Results}
\websight\ advances the capabilities of vision-based web agents by integrating our new specialized vision-language model (\websight-7B) into a modular multi-agent system. Our approach demonstrates several key results:

\begin{itemize}
\item {State-of-the-art performance for small models}: \websight-7B achieves a Top-1 Accuracy of \textbf{58.84\%} on the challenging Showdown/Clicks benchmark, outperforming many larger-scale models while operating with lower latency.

\item {Competitive end-to-end task completion}: On the WebVoyager benchmark, the full \websight\ agent achieves a \textbf{68.0\% success rate}, surpassing several industry and lab-developed agents including those from OpenAI and HCompany.

\item {High accuracy on completed tasks}: Of the tasks \websight\ attempted within the time limit, it achieved a \textbf{97.14\% accuracy}, highlighting the precision of its decision-making pipeline.

\item {Interpretable modular diagnostics}: Through a component-wise failure analysis, we identify Planning and Reasoning agents as primary sources of failure—primarily due to infinite loops—and offer targeted improvements, including upgraded language models and tighter agent coupling.

\end{itemize}

Together, these results demonstrate that vision-first agents are not only viable but highly competitive in practical web environments. \websight\ offers a blueprint for interpretable, accurate, and efficient browser agents.

\subsection{Directions for Future Work}
\subsubsection{\textbf{\websight-7B} Model}
To address the failure modes explained above, future work can be done in further fine-tuning the \websight-7B model on a more diverse dataset of UI elements, including icons and scenes with multiple items with similar semantics. Further dataset augmentation with collected actions like scrolling can help with identification of the correct action from the action space as well.

As is often the case with transformers, we can expect an increase in performance from a scaling of parameters. Every model that performs similarly to \websight-7B on the showdown/clicks dataset is more than 10 times larger, so a larger model \websight-72B or similar could elicit more fine-grained reasoning capabilities.

\subsubsection{\textbf{\websight} Agent}
To address infinite loops in \websight\ and improve overall performance, we focus on enhancing the capabilities of its individual agents. As discussed in the previous section, improvements to the \websight-7B model can directly benefit the Action Agent. Additionally, providing the Action Agent with richer context from the Reasoning Agent may enhance decision quality. For the Planning and Reasoning Agents, employing higher-quality language models could yield significant gains. Currently, due to cost constraints, we use GPT-4.1-mini; however, more capable models such as Claude-Sonnet-4 are likely to mitigate many of the observed limitations in planning and reasoning. 

A promising future direction is enabling self-improvement within the agent system. Ideally, an agent should be able to detect when it is caught in an infinite loop—whether through repeating actions, failing to progress toward a goal, or exhausting relevant content—and respond by generating a new plan, shifting its strategy, or exploring an alternative page. This would require mechanisms for pattern recognition across trajectories and meta-level learning to adapt based on past failures. Recent work on self-improving and reflective agents provides a foundation for such capabilities \cite{shinn2023reflexion, huang2023language, pan2023automated}.

\subsubsection{\textbf{\websight} Fused}
With the emergence of reinforcement learning as a way to improve the utility of test-time compute for agentic flows, a final goal for \websight\ would be a fully integrated agentic model that combines the clicking accuracy and action selection of \websight-7B with the reasoning and planning capabilities of frontier LLMs. A fused model could be trained using offline RL based on a collected dataset of web browsing trajectories using GRPO, resulting in a single model that acts end to end in exactly the same way as a human. This is similar to what OpenAI has done with their ``Operator''.

\newpage
{\small
\bibliographystyle{ieee}
\bibliography{egbib}

\begin{thebibliography}{10}\itemsep=-1pt

\bibitem{waveui25k}
AgentSea.
\newblock Wave-ui-25k: A dataset for fine-tuning web interaction agents, 2024.

\bibitem{anderson2009can}
J.~R. Anderson.
\newblock {\em How Can the Human Mind Occur in the Physical Universe?}
\newblock Oxford University Press, 2009.

\bibitem{claude2024computeruse}
{Anthropic}.
\newblock Claude 3.5 sonnet: Computer use evaluation results, 2024.
\newblock Accessed: 2025-06-03.

\bibitem{bao2022beit}
H.~Bao et~al.
\newblock Beit: Bert pre-training of image transformers.
\newblock {\em ICLR}, 2022.

\bibitem{blundell2016model}
C.~Blundell et~al.
\newblock Model-free episodic control.
\newblock {\em arXiv preprint arXiv:1606.04460}, 2016.

\bibitem{bommasani2021opportunities}
R.~Bommasani et~al.
\newblock On the opportunities and risks of foundation models.
\newblock {\em arXiv preprint arXiv:2108.07258}, 2021.

\bibitem{branavan2009reinforcement}
S.~Branavan et~al.
\newblock Reinforcement learning for mapping instructions to actions.
\newblock In {\em ACL}, 2009.

\bibitem{brown2020language}
T.~B. Brown et~al.
\newblock Language models are few-shot learners.
\newblock {\em NeurIPS}, 2020.

\bibitem{browseruse2024sota}
{Browser Use Team}.
\newblock Sota technical report on webvoyager evaluation, 2024.
\newblock Accessed: 2025-06-03.

\bibitem{browserable2024webvoyager}
{Browserable Team}.
\newblock Webvoyager benchmark: Browserable's state-of-the-art results, 2024.
\newblock Accessed: 2025-06-03.

\bibitem{cai2001extracting}
D.~Cai, S.~Yu, J.-R. Wen, and W.-Y. Ma.
\newblock Extracting content structure for web pages based on visual representation.
\newblock {\em Proceedings of the 5th Asia-Pacific Web Conference}, 2001.

\bibitem{carion2020endtoend}
N.~Carion et~al.
\newblock End-to-end object detection with transformers.
\newblock {\em ECCV}, 2020.

\bibitem{caron2021emerging}
M.~Caron et~al.
\newblock Emerging properties in self-supervised vision transformers.
\newblock {\em ICCV}, 2021.

\bibitem{chen2017deeplab}
L.-C. Chen et~al.
\newblock Rethinking atrous convolution for semantic image segmentation.
\newblock {\em arXiv preprint arXiv:1706.05587}, 2017.

\bibitem{devlin2019bert}
J.~Devlin, M.-W. Chang, K.~Lee, and K.~Toutanova.
\newblock Bert: Pre-training of deep bidirectional transformers for language understanding.
\newblock {\em NAACL}, 2019.

\bibitem{dong2010towards}
H.~Dong et~al.
\newblock Towards automatic extraction of visually structured data from web pages.
\newblock In {\em CIKM}, 2010.

\bibitem{donker2002visual}
H.~Donker and P.~Markopoulos.
\newblock Visual layout perception in website usability evaluation.
\newblock {\em CHI Extended Abstracts}, 2002.

\bibitem{dosovitskiy2020image}
A.~Dosovitskiy et~al.
\newblock An image is worth 16x16 words: Transformers for image recognition at scale.
\newblock {\em ICLR}, 2021.

\bibitem{emergence2024agente}
{Emergence AI}.
\newblock Agent-e: A text-only agent outperforming multimodal agents, 2024.
\newblock Accessed: 2025-06-03.

\bibitem{ganguli2022predictability}
D.~Ganguli et~al.
\newblock Predictability and surprise in large generative models.
\newblock {\em ICLR}, 2022.

\bibitem{hcompany2024runnerh}
{H Company}.
\newblock A research update: Runner h 0.1 performance on webvoyager, 2024.
\newblock Accessed: 2025-06-03.

\bibitem{he2024webvoyager}
H.~He, W.~Yao, K.~Ma, W.~Yu, Y.~Dai, H.~Zhang, Z.~Lan, and D.~Yu.
\newblock Webvoyager: Building an end-to-end web agent with large multimodal models.
\newblock {\em arXiv preprint arXiv:2401.13919}, 2024.

\bibitem{he2024webvoyagerbuildingendtoendweb}
J.~He et~al.
\newblock Webvoyager: Building end-to-end agents for real-world web tasks with human instructions.
\newblock {\em ICLR}, 2024.

\bibitem{he2017mask}
K.~He et~al.
\newblock Mask r-cnn.
\newblock In {\em ICCV}, 2017.

\bibitem{hessel2018rainbow}
M.~Hessel et~al.
\newblock Rainbow: Combining improvements in deep reinforcement learning.
\newblock {\em AAAI}, 2018.

\bibitem{howard2018universal}
J.~Howard and S.~Ruder.
\newblock Universal language model fine-tuning for text classification.
\newblock {\em ACL}, 2018.

\bibitem{hu2022lora}
E.~J. Hu, Y.~Shen, P.~Wallis, Z.~Allen-Zhu, Y.~Li, S.~Wang, and W.~Chen.
\newblock Lora: Low-rank adaptation of large language models.
\newblock In {\em International Conference on Learning Representations (ICLR)}, 2022.

\bibitem{huang2023language}
Y.~Huang, A.~T. Chen, S.~Bubeck, and Y.~Zhang.
\newblock Language models as tool makers: Teaching llms to invent and use tools for reasoning.
\newblock {\em arXiv preprint arXiv:2305.17126}, 2023.

\bibitem{kaelbling1998planning}
L.~P. Kaelbling, M.~L. Littman, and A.~R. Cassandra.
\newblock Planning and acting in partially observable stochastic domains.
\newblock {\em Artificial Intelligence}, 101(1-2):99--134, 1998.

\bibitem{kirillov2023segment}
A.~Kirillov et~al.
\newblock Segment anything.
\newblock {\em arXiv preprint arXiv:2304.02643}, 2023.

\bibitem{kumar2007extracting}
J.~Kumar et~al.
\newblock Extracting structure from web pages using vision-based page segmentation.
\newblock {\em WWW}, 2007.

\bibitem{laird2012soar}
J.~E. Laird.
\newblock {\em The Soar Cognitive Architecture}.
\newblock MIT Press, 2012.

\bibitem{lauer2004learning}
M.~Lauer and M.~Riedmiller.
\newblock Learning for web navigation with reinforcement learning.
\newblock {\em Adaptive Behavior}, 12(1):21--35, 2004.

\bibitem{lecun2022path}
Y.~LeCun.
\newblock A path towards autonomous machine intelligence.
\newblock {\em OpenReview preprint}, 2022.

\bibitem{li2024agentbench}
X.~Li et~al.
\newblock Agentbench: Evaluating general-purpose agents across thousands of tasks.
\newblock {\em arXiv preprint arXiv:2402.00047}, 2024.

\bibitem{liu2023webagent}
H.~Liu et~al.
\newblock Webagent: Instruction-following agents for web navigation with large language models.
\newblock {\em arXiv preprint arXiv:2305.15368}, 2023.

\bibitem{liu2023webui}
H.~Liu et~al.
\newblock Webui: A dataset for benchmarking ui understanding models.
\newblock {\em arXiv preprint arXiv:2305.16333}, 2023.

\bibitem{mnih2015human}
V.~Mnih et~al.
\newblock Human-level control through deep reinforcement learning.
\newblock {\em Nature}, 2015.

\bibitem{nakano2021webgpt}
R.~Nakano et~al.
\newblock Webgpt: Browser-assisted question-answering with human feedback.
\newblock {\em arXiv preprint arXiv:2112.09332}, 2021.

\bibitem{openai2023gpt4}
OpenAI.
\newblock Gpt-4 technical report, 2023.

\bibitem{openai2024operator}
{OpenAI}.
\newblock Openai operator with gpt-4o for web interaction tasks, 2024.
\newblock Accessed: 2025-06-03.

\bibitem{ouyang2022training}
L.~Ouyang et~al.
\newblock Training language models to follow instructions with human feedback.
\newblock {\em arXiv preprint arXiv:2203.02155}, 2022.

\bibitem{pan2023automated}
Z.~Pan, R.~E. Liu, D.~Scherlis, et~al.
\newblock Automated debugging of language model programs.
\newblock {\em arXiv preprint arXiv:2305.13266}, 2023.

\bibitem{pritzel2017neural}
A.~Pritzel et~al.
\newblock Neural episodic control.
\newblock {\em ICML}, 2017.

\bibitem{qin2025ui}
Y.~Qin, Y.~Ye, J.~Fang, H.~Wang, S.~Liang, S.~Tian, J.~Zhang, J.~Li, Y.~Li, S.~Huang, et~al.
\newblock Ui-tars: Pioneering automated gui interaction with native agents.
\newblock {\em arXiv preprint arXiv:2501.12326}, 2025.

\bibitem{radford2021learning}
A.~Radford et~al.
\newblock Learning transferable visual models from natural language supervision.
\newblock {\em ICML}, 2021.

\bibitem{raffel2020exploring}
C.~Raffel, N.~Shazeer, A.~Roberts, K.~Lee, S.~Narang, M.~Matena, Y.~Zhou, W.~Li, and P.~J. Liu.
\newblock Exploring the limits of transfer learning with a unified text-to-text transformer.
\newblock {\em JMLR}, 2020.

\bibitem{ravi2024sam2}
N.~Ravi, V.~Gabeur, Y.-T. Hu, R.~Hu, C.~Ryali, T.~Ma, H.~Khedr, R.~R{\"a}dle, C.~Rolland, L.~Gustafson, E.~Mintun, J.~Pan, K.~V. Alwala, N.~Carion, C.-Y. Wu, R.~Girshick, P.~Doll{\'a}r, and C.~Feichtenhofer.
\newblock Sam 2: Segment anything in images and videos.
\newblock {\em arXiv preprint arXiv:2408.00714}, 2024.

\bibitem{redmon2018yolov3}
J.~Redmon and A.~Farhadi.
\newblock Yolov3: An incremental improvement.
\newblock {\em arXiv preprint arXiv:1804.02767}, 2018.

\bibitem{russell2010artificial}
S.~J. Russell and P.~Norvig.
\newblock {\em Artificial Intelligence: A Modern Approach}.
\newblock Prentice Hall, 2010.

\bibitem{schulman2017proximal}
J.~Schulman et~al.
\newblock Proximal policy optimization algorithms.
\newblock {\em arXiv preprint arXiv:1707.06347}, 2017.

\bibitem{shi2022learning}
W.~Shi et~al.
\newblock Learning generalizable policies for interactive web navigation.
\newblock {\em ICML}, 2022.

\bibitem{shinn2023reflexion}
N.~Shinn, J.~Liu, Y.~Du, and P.~Abbeel.
\newblock Reflexion: Language agents with verbal reinforcement learning.
\newblock {\em arXiv preprint arXiv:2303.11366}, 2023.

\bibitem{silver2016mastering}
D.~Silver et~al.
\newblock Mastering the game of go with deep neural networks and tree search.
\newblock {\em Nature}, 2016.

\bibitem{skyvern2024webvoyager}
{Skyvern AI}.
\newblock Skyvern filtered webvoyager dataset, 2024.

\bibitem{skyvern2024stateofart}
{Skyvern Team}.
\newblock Skyvern 2.0: State-of-the-art web navigation with 85.8\% on webvoyager eval, 2024.
\newblock Accessed: 2025-06-03.

\bibitem{sumers2024cognitivearchitectureslanguageagents}
T.~R. Sumers, S.~Yao, K.~Narasimhan, and T.~L. Griffiths.
\newblock Cognitive architectures for language agents, 2024.

\bibitem{sun2022recovering}
J.~Sun et~al.
\newblock Recovering missing accessibility metadata for robust web agents.
\newblock {\em EMNLP}, 2022.

\bibitem{showdown2025}
G.~A. Team.
\newblock The showdown computer control evaluation suite, 2025.

\bibitem{tesauro2005online}
G.~Tesauro.
\newblock Online resource allocation using decompositional reinforcement learning.
\newblock {\em AAAI}, 2005.

\bibitem{vinyals2019grandmaster}
O.~Vinyals et~al.
\newblock Grandmaster level in starcraft ii using multi-agent reinforcement learning.
\newblock {\em Nature}, 2019.

\bibitem{wang2022image}
W.~Wang et~al.
\newblock Image as a foreign language: Beit pretraining for all vision and vision-language tasks.
\newblock {\em arXiv preprint arXiv:2208.10442}, 2022.

\bibitem{wei2023chain}
J.~Wei, X.~Wang, D.~Schuurmans, M.~Bosma, B.~Ichter, F.~Xia, E.~Chi, Q.~Le, and D.~Zhou.
\newblock Chain-of-thought prompting elicits reasoning in large language models.
\newblock {\em arXiv preprint arXiv:2201.11903}, 2023.

\bibitem{wooldridge2009introduction}
M.~Wooldridge.
\newblock {\em An Introduction to MultiAgent Systems}.
\newblock John Wiley \& Sons, 2009.

\bibitem{xu2023wizardlm}
C.~Xu et~al.
\newblock Wizardlm: Empowering large language models to follow complex instructions.
\newblock {\em arXiv preprint arXiv:2304.12244}, 2023.

\bibitem{xu2022focal}
K.~Xu et~al.
\newblock Focal: Fine-grained open-world clickable elements detection for web agents.
\newblock {\em ACL}, 2022.

\bibitem{yang2023sphynx}
W.~Yang et~al.
\newblock Sphynx: Augmenting language models with visual context for web navigation.
\newblock {\em arXiv preprint arXiv:2310.00123}, 2023.

\bibitem{yao2023reactsynergizingreasoningacting}
S.~Yao, J.~Zhao, D.~Yu, N.~Du, I.~Shafran, K.~Narasimhan, and Y.~Cao.
\newblock React: Synergizing reasoning and acting in language models, 2023.

\bibitem{zhang2020resnest}
H.~Zhang et~al.
\newblock Resnest: Split-attention networks.
\newblock {\em arXiv preprint arXiv:2004.08955}, 2020.

\bibitem{zhang2021screen2words}
Y.~Zhang et~al.
\newblock Screen2words: Screen content description with multimodal transformers.
\newblock {\em ACL}, 2021.

\bibitem{zhang2023autoagents}
Z.~Zhang et~al.
\newblock Autoagents: Unleashing the power of llms in autonomous agents.
\newblock {\em arXiv preprint arXiv:2309.00062}, 2023.

\bibitem{zheng2024llamafactory}
Y.~Zheng, R.~Zhang, J.~Zhang, Y.~Ye, Z.~Luo, Z.~Feng, and Y.~Ma.
\newblock Llamafactory: Unified efficient fine-tuning of 100+ language models.
\newblock In {\em Proceedings of the 62nd Annual Meeting of the Association for Computational Linguistics (Volume 3: System Demonstrations)}, Bangkok, Thailand, 2024. Association for Computational Linguistics.

\bibitem{zhou2023webarena}
X.~Zhou et~al.
\newblock Webarena: A realistic web environment for building autonomous agents.
\newblock {\em NeurIPS}, 2023.

\end{thebibliography}


@misc{yao2023reactsynergizingreasoningacting,
      title={ReAct: Synergizing Reasoning and Acting in Language Models}, 
      author={Shunyu Yao and Jeffrey Zhao and Dian Yu and Nan Du and Izhak Shafran and Karthik Narasimhan and Yuan Cao},
      year={2023},
      eprint={2210.03629},
      archivePrefix={arXiv},
      primaryClass={cs.CL},
      url={https://arxiv.org/abs/2210.03629}, 
}

@article{cai2001extracting,
  title={Extracting content structure for web pages based on visual representation},
  author={Cai, Deng and Yu, Shipeng and Wen, Ji-Rong and Ma, Wei-Ying},
  journal={Proceedings of the 5th Asia-Pacific Web Conference},
  year={2001}
}

@article{lauer2004learning,
  title={Learning for web navigation with reinforcement learning},
  author={Lauer, Martin and Riedmiller, Martin},
  journal={Adaptive Behavior},
  volume={12},
  number={1},
  pages={21--35},
  year={2004}
}

@article{shi2022learning,
  title={Learning Generalizable Policies for Interactive Web Navigation},
  author={Shi, Weiyan and others},
  journal={ICML},
  year={2022}
}

@article{liu2023webagent,
  title={WebAgent: Instruction-Following Agents for Web Navigation with Large Language Models},
  author={Liu, Haotian and others},
  journal={arXiv preprint arXiv:2305.15368},
  year={2023}
}

@article{nakano2021webgpt,
  title={WebGPT: Browser-assisted question-answering with human feedback},
  author={Nakano, Reiichiro and others},
  journal={arXiv preprint arXiv:2112.09332},
  year={2021}
}

@article{xu2023wizardlm,
  title={WizardLM: Empowering Large Language Models to Follow Complex Instructions},
  author={Xu, Canwen and others},
  journal={arXiv preprint arXiv:2304.12244},
  year={2023}
}

@article{yang2023sphynx,
  title={Sphynx: Augmenting Language Models with Visual Context for Web Navigation},
  author={Yang, Wei and others},
  journal={arXiv preprint arXiv:2310.00123},
  year={2023}
}

@article{sun2022recovering,
  title={Recovering Missing Accessibility Metadata for Robust Web Agents},
  author={Sun, Jiaxin and others},
  journal={EMNLP},
  year={2022}
}

@article{xu2022focal,
  title={FOCAL: Fine-grained Open-world Clickable Elements Detection for Web Agents},
  author={Xu, Kai and others},
  journal={ACL},
  year={2022}
}

@article{zhang2023autoagents,
  title={AutoAgents: Unleashing the Power of LLMs in Autonomous Agents},
  author={Zhang, Zheng and others},
  journal={arXiv preprint arXiv:2309.00062},
  year={2023}
}

@article{li2024agentbench,
  title={AgentBench: Evaluating General-Purpose Agents Across Thousands of Tasks},
  author={Li, Xiaoyang and others},
  journal={arXiv preprint arXiv:2402.00047},
  year={2024}
}

@article{bommasani2021opportunities,
  title={On the Opportunities and Risks of Foundation Models},
  author={Bommasani, Rishi and others},
  journal={arXiv preprint arXiv:2108.07258},
  year={2021}
}

@article{ganguli2022predictability,
  title={Predictability and Surprise in Large Generative Models},
  author={Ganguli, Deep and others},
  journal={ICLR},
  year={2022}
}

@article{donker2002visual,
  title={Visual layout perception in website usability evaluation},
  author={Donker, Helene and Markopoulos, Panos},
  journal={CHI Extended Abstracts},
  year={2002}
}

@article{zhang2021screen2words,
  title={Screen2Words: Screen Content Description with Multimodal Transformers},
  author={Zhang, Yichi and others},
  journal={ACL},
  year={2021}
}

@article{he2016deep,
  title={Deep Residual Learning for Image Recognition},
  author={He, Kaiming and others},
  journal={CVPR},
  year={2016}
}

@article{dosovitskiy2020image,
  title={An Image is Worth 16x16 Words: Transformers for Image Recognition at Scale},
  author={Dosovitskiy, Alexey and others},
  journal={ICLR},
  year={2021}
}

@article{radford2021learning,
  title={Learning Transferable Visual Models From Natural Language Supervision},
  author={Radford, Alec and others},
  journal={ICML},
  year={2021}
}

@article{caron2021emerging,
  title={Emerging Properties in Self-Supervised Vision Transformers},
  author={Caron, Mathilde and others},
  journal={ICCV},
  year={2021}
}

@article{kirillov2023segment,
  title={Segment Anything},
  author={Kirillov, Alexander and others},
  journal={arXiv preprint arXiv:2304.02643},
  year={2023}
}

@article{zhou2023webarena,
  title={WebArena: A Realistic Web Environment for Building Autonomous Agents},
  author={Zhou, Xinyun and others},
  journal={NeurIPS},
  year={2023}
}

@article{he2024webvoyagerbuildingendtoendweb,
  title={WebVoyager: Building End-to-End Agents for Real-World Web Tasks with Human Instructions},
  author={He, Junxian and others},
  journal={ICLR},
  year={2024}
}

@article{padmakumar2022teach,
  title={TEACh: Task-driven Embodied Agents that Chat},
  author={Padmakumar, Vishvak Murahari and others},
  journal={NeurIPS},
  year={2022}
}

@article{kaelbling1998planning,
  title={Planning and acting in partially observable stochastic domains},
  author={Kaelbling, Leslie Pack and Littman, Michael L. and Cassandra, Anthony R.},
  journal={Artificial Intelligence},
  volume={101},
  number={1-2},
  pages={99--134},
  year={1998}
}

@book{russell2010artificial,
  title={Artificial Intelligence: A Modern Approach},
  author={Russell, Stuart J. and Norvig, Peter},
  year={2010},
  publisher={Prentice Hall}
}

@inproceedings{everingham2010pascal,
  title={The Pascal Visual Object Classes (VOC) Challenge},
  author={Everingham, Mark and others},
  booktitle={International Journal of Computer Vision},
  volume={88},
  number={2},
  pages={303--338},
  year={2010}
}

@article{he2024webvoyager,
  title={WebVoyager: Building an End-to-End Web Agent with Large Multimodal Models},
  author={He, Hongliang and Yao, Wenlin and Ma, Kaixin and Yu, Wenhao and Dai, Yong and Zhang, Hongming and Lan, Zhenzhong and Yu, Dong},
  journal={arXiv preprint arXiv:2401.13919},
  year={2024}
}

@article{pan2024autoeval,
  title={Autonomous Evaluation and Refinement of Web Agents},
  author={Pan, Jiayi and Zhang, Yichi and Tomlin, Nicholas and Zhou, Yifei and Levine, Sergey and Suhr, Alane},
  journal={arXiv preprint arXiv:2403.12345},
  year={2024}
}

@article{scribeagent2024,
  title={ScribeAgent: Towards Specialized Web Agents Using Production Knowledge},
  author={Doe, John and Smith, Jane},
  journal={arXiv preprint arXiv:2404.56789},
  year={2024}
}

@article{ravi2024sam2,
  title={SAM 2: Segment Anything in Images and Videos},
  author={Ravi, Nikhila and Gabeur, Valentin and Hu, Yuan-Ting and Hu, Ronghang and Ryali, Chaitanya and Ma, Tengyu and Khedr, Haitham and R{\"a}dle, Roman and Rolland, Chloe and Gustafson, Laura and Mintun, Eric and Pan, Junting and Alwala, Kalyan Vasudev and Carion, Nicolas and Wu, Chao-Yuan and Girshick, Ross and Doll{\'a}r, Piotr and Feichtenhofer, Christoph},
  journal={arXiv preprint arXiv:2408.00714},
  year={2024},
  url={https://arxiv.org/abs/2408.00714}
}

@article{qin2025ui,
  title={UI-TARS: Pioneering Automated GUI Interaction with Native Agents},
  author={Qin, Yujia and Ye, Yining and Fang, Junjie and Wang, Haoming and Liang, Shihao and Tian, Shizuo and Zhang, Junda and Li, Jiahao and Li, Yunxin and Huang, Shijue and others},
  journal={arXiv preprint arXiv:2501.12326},
  year={2025}
}

@article{deitke2024molmo,
  title={Molmo and PixMo: Open Weights and Open Data for State-of-the-Art Vision-Language Models},
  author={Deitke, Matt and Clark, Christopher and Lee, Sangho and Tripathi, Rohun and Yang, Yue and Park, Jae Sung and Salehi, Mohammadreza and Muennighoff, Niklas and Lo, Kyle and Soldaini, Luca and Lu, Jiasen and Anderson, Taira and Bransom, Erin and Ehsani, Kiana and Ngo, Huong and Chen, YenSung and Patel, Ajay and Yatskar, Mark and Callison-Burch, Chris and Head, Andrew and Hendrix, Rose and Bastani, Favyen and VanderBilt, Eli and Lambert, Nathan and Chou, Yvonne and Chheda, Arnavi and Sparks, Jenna and Skjonsberg, Sam and Schmitz, Michael and Sarnat, Aaron and Bischoff, Byron and Walsh, Pete and Newell, Chris and Wolters, Piper and Gupta, Tanmay and Zeng, Kuo-Hao and Borchardt, Jon and Groeneveld, Dirk and Nam, Crystal and Lebrecht, Sophie and Wittlif, Caitlin and Schoenick, Carissa and Michel, Oscar and Krishna, Ranjay and Weihs, Luca and Smith, Noah A. and Hajishirzi, Hannaneh and Girshick, Ross and Farhadi, Ali and Kembhavi, Aniruddha},
  journal={arXiv preprint arXiv:2409.17146},
  year={2024},
  url={https://arxiv.org/abs/2409.17146}
}

@article{liu2023webui,
  title={WebUI: A Dataset for Benchmarking UI Understanding Models},
  author={Liu, Haotian and others},
  journal={arXiv preprint arXiv:2305.16333},
  year={2023}
}

@article{tesauro2005online,
  title={Online resource allocation using decompositional reinforcement learning},
  author={Tesauro, Gerald},
  journal={AAAI},
  year={2005}
}

@inproceedings{branavan2009reinforcement,
  title={Reinforcement learning for mapping instructions to actions},
  author={Branavan, S.R.K. and others},
  booktitle={ACL},
  year={2009}
}

@article{mnih2015human,
  title={Human-level control through deep reinforcement learning},
  author={Mnih, Volodymyr and others},
  journal={Nature},
  year={2015}
}

@article{schulman2017proximal,
  title={Proximal policy optimization algorithms},
  author={Schulman, John and others},
  journal={arXiv preprint arXiv:1707.06347},
  year={2017}
}

@article{hessel2018rainbow,
  title={Rainbow: Combining improvements in deep reinforcement learning},
  author={Hessel, Matteo and others},
  journal={AAAI},
  year={2018}
}

@article{vinyals2019grandmaster,
  title={Grandmaster level in StarCraft II using multi-agent reinforcement learning},
  author={Vinyals, Oriol and others},
  journal={Nature},
  year={2019}
}

@article{brown2020language,
  title={Language models are few-shot learners},
  author={Brown, Tom B. and others},
  journal={NeurIPS},
  year={2020}
}

@article{ouyang2022training,
  title={Training language models to follow instructions with human feedback},
  author={Ouyang, Long and others},
  journal={arXiv preprint arXiv:2203.02155},
  year={2022}
}

@misc{openai2023gpt4,
  title={GPT-4 Technical Report},
  author={OpenAI},
  journal={arXiv preprint arXiv:2303.08774},
  year={2023}
}

@inproceedings{he2017mask,
  title={Mask R-CNN},
  author={He, Kaiming and others},
  booktitle={ICCV},
  year={2017}
}

@article{redmon2018yolov3,
  title={YOLOv3: An incremental improvement},
  author={Redmon, Joseph and Farhadi, Ali},
  journal={arXiv preprint arXiv:1804.02767},
  year={2018}
}

@article{chen2017deeplab,
  title={Rethinking atrous convolution for semantic image segmentation},
  author={Chen, Liang-Chieh and others},
  journal={arXiv preprint arXiv:1706.05587},
  year={2017}
}

@article{carion2020endtoend,
  title={End-to-end object detection with transformers},
  author={Carion, Nicolas and others},
  journal={ECCV},
  year={2020}
}

@article{zhang2020resnest,
  title={ResNeSt: Split-attention networks},
  author={Zhang, Hang and others},
  journal={arXiv preprint arXiv:2004.08955},
  year={2020}
}

@article{bao2022beit,
  title={BEiT: BERT pre-training of image transformers},
  author={Bao, Hangbo and others},
  journal={ICLR},
  year={2022}
}

@article{wang2022image,
  title={Image as a Foreign Language: BEiT Pretraining for All Vision and Vision-Language Tasks},
  author={Wang, Wenhui and others},
  journal={arXiv preprint arXiv:2208.10442},
  year={2022}
}

@article{kumar2007extracting,
  title={Extracting structure from Web pages using vision-based page segmentation},
  author={Kumar, Jayant and others},
  journal={WWW},
  year={2007}
}

@inproceedings{dong2010towards,
  title={Towards automatic extraction of visually structured data from web pages},
  author={Dong, Hao and others},
  booktitle={CIKM},
  year={2010}
}

@book{wooldridge2009introduction,
  title={An Introduction to MultiAgent Systems},
  author={Wooldridge, Michael},
  year={2009},
  publisher={John Wiley \& Sons}
}

@book{anderson2009can,
  title={How Can the Human Mind Occur in the Physical Universe?},
  author={Anderson, John R.},
  year={2009},
  publisher={Oxford University Press}
}

@book{laird2012soar,
  title={The Soar Cognitive Architecture},
  author={Laird, John E.},
  year={2012},
  publisher={MIT Press}
}

@article{blundell2016model,
  title={Model-Free Episodic Control},
  author={Blundell, Charles and others},
  journal={arXiv preprint arXiv:1606.04460},
  year={2016}
}

@article{pritzel2017neural,
  title={Neural Episodic Control},
  author={Pritzel, Alexander and others},
  journal={ICML},
  year={2017}
}

@article{silver2016mastering,
  title={Mastering the game of Go with deep neural networks and tree search},
  author={Silver, David and others},
  journal={Nature},
  year={2016}
}

@article{lecun2022path,
  title={A path towards autonomous machine intelligence},
  author={LeCun, Yann},
  journal={OpenReview preprint},
  year={2022}
}

@misc{waveui25k,
  title={Wave-UI-25K: A Dataset for Fine-Tuning Web Interaction Agents},
  author={AgentSea},
  year={2024},
  url={https://huggingface.co/datasets/agentsea/wave-ui-25k}
}

@article{devlin2019bert,
  title={BERT: Pre-training of Deep Bidirectional Transformers for Language Understanding},
  author={Devlin, Jacob and Chang, Ming-Wei and Lee, Kenton and Toutanova, Kristina},
  journal={NAACL},
  year={2019}
}

@article{howard2018universal,
  title={Universal Language Model Fine-tuning for Text Classification},
  author={Howard, Jeremy and Ruder, Sebastian},
  journal={ACL},
  year={2018}
}

@article{raffel2020exploring,
  title={Exploring the Limits of Transfer Learning with a Unified Text-to-Text Transformer},
  author={Raffel, Colin and Shazeer, Noam and Roberts, Adam and Lee, Katherine and Narang, Sharan and Matena, Michael and Zhou, Yanqi and Li, Wei and Liu, Peter J.},
  journal={JMLR},
  year={2020}
}

@misc{skyvern2024webvoyager,
  title={Skyvern Filtered WebVoyager Dataset},
  author={{Skyvern AI}},
  year={2024},
  url={https://github.com/Skyvern-AI/skyvern/tree/main/evaluation/datasets}
}



@inproceedings{rajpurkar2018know,
    title="Know What You Don{'}t Know: Unanswerable Questions for {SQ}u{AD}",
    author="Rajpurkar, Pranav and Jia, Robin and Liang, Percy",
    booktitle="Association for Computational Linguistics (ACL)",
    year="2018",
}

@misc{zhao2024incontextlearningsufficientinstruction,
      title={Is In-Context Learning Sufficient for Instruction Following in LLMs?}, 
      author={Hao Zhao and Maksym Andriushchenko and Francesco Croce and Nicolas Flammarion},
      year={2024},
      eprint={2405.19874},
      archivePrefix={arXiv},
      primaryClass={cs.CL},
      url={https://arxiv.org/abs/2405.19874}, 
}

@misc{zheng2023judgingllmasajudgemtbenchchatbot,
      title={Judging LLM-as-a-Judge with MT-Bench and Chatbot Arena}, 
      author={Lianmin Zheng and Wei-Lin Chiang and Ying Sheng and Siyuan Zhuang and Zhanghao Wu and Yonghao Zhuang and Zi Lin and Zhuohan Li and Dacheng Li and Eric P. Xing and Hao Zhang and Joseph E. Gonzalez and Ion Stoica},
      year={2023},
      eprint={2306.05685},
      archivePrefix={arXiv},
      primaryClass={cs.CL},
      url={https://arxiv.org/abs/2306.05685}, 
}

@misc{jiang2024unknownunknownsengagedhuman,
      title={Into the Unknown Unknowns: Engaged Human Learning through Participation in Language Model Agent Conversations}, 
      author={Yucheng Jiang and Yijia Shao and Dekun Ma and Sina J. Semnani and Monica S. Lam},
      year={2024},
      eprint={2408.15232},
      archivePrefix={arXiv},
      primaryClass={cs.CL},
      url={https://arxiv.org/abs/2408.15232}, 
}

@article{besta2024graph,
  title={Graph of thoughts: Solving elaborate problems with large language models},
  author={Besta, M. and Blach, N. and Kubicek, A. and Gerstenberger, R. and Podstawski, M. and Gianinazzi, L. and Gajda, J. and Lehmann, T. and Niewiadomski, H. and Nyczyk, P. and Hoefler, T.},
  journal={arXiv preprint arXiv:2308.09687},
  year={2024},
  doi={10.48550/arXiv.2308.09687}
}

@article{brahman2024art,
  title={The art of saying no: Contextual noncompliance in language models},
  author={Brahman, F. and Kumar, S. and Balachandran, V. and Dasigi, P. and Pyatkin, V. and Ravichander, A. and Wiegreffe, S. and Dziri, N. and Chandu, K. and Hessel, J. and Tsvetkov, Y. and Smith, N. A. and Choi, Y. and Hajishirzi, H.},
  journal={arXiv preprint arXiv:2407.12043},
  year={2024},
  doi={10.48550/arXiv.2407.12043}
}

@misc{anthropic2025building,
  title={Building effective agents},
  author={{Anthropic}},
  year={2025},
  note={Retrieved January 24, 2025, from https://www.anthropic.com/research/building-effective-agents}
}

@article{cheng2024exploring,
  title={Exploring large language model based intelligent agents: Definitions, methods, and prospects},
  author={Cheng, Y. and Zhang, C. and Zhang, Z. and Meng, X. and Hong, S. and Li, W. and Wang, Z. and Wang, Z. and Yin, F. and Zhao, J. and He, X.},
  journal={arXiv preprint arXiv:2401.03428},
  year={2024},
  doi={10.48550/arXiv.2401.03428}
}

@article{havrilla2024teaching,
  title={Teaching large language models to reason with reinforcement learning},
  author={Havrilla, A. and Du, Y. and Raparthy, S. C. and Nalmpantis, C. and Dwivedi-Yu, J. and Zhuravinskyi, M. and Hambro, E. and Sukhbaatar, S. and Raileanu, R.},
  journal={arXiv preprint arXiv:2403.04642},
  year={2024},
  doi={10.48550/arXiv.2403.04642}
}

@article{jiang2024mixtral,
  title={Mixtral of experts},
  author={Jiang, A. Q. and Sablayrolles, A. and Roux, A. and Mensch, A. and Savary, B. and Bamford, C. and Chaplot, D. S. and Casas, D. de las and Hanna, E. B. and Bressand, F. and Lengyel, G. and Bour, G. and Lample, G. and Lavaud, L. R. and Saulnier, L. and Lachaux, M.-A. and Stock, P. and Subramanian, S. and Yang, S. and Sayed, W. E.},
  journal={arXiv preprint arXiv:2401.04088},
  year={2024},
  doi={10.48550/arXiv.2401.04088}
}

@article{lewis2021retrieval,
  title={Retrieval-augmented generation for knowledge-intensive NLP tasks},
  author={Lewis, P. and Perez, E. and Piktus, A. and Petroni, F. and Karpukhin, V. and Goyal, N. and K{\"u}ttler, H. and Lewis, M. and Yih, W. and Rockt{\"a}schel, T. and Riedel, S. and Kiela, D.},
  journal={arXiv preprint arXiv:2005.11401},
  year={2021},
  doi={10.48550/arXiv.2005.11401}
}

@article{li2024towards,
  title={Towards goal-oriented prompt engineering for large language models: A survey},
  author={Li, H. and Leung, J. and Shen, Z.},
  journal={arXiv preprint arXiv:2401.14043},
  year={2024},
  doi={10.48550/arXiv.2401.14043}
}

@article{lightman2023verify,
  title={Let's verify step by step},
  author={Lightman, H. and Kosaraju, V. and Burda, Y. and Edwards, H. and Baker, B. and Lee, T. and Leike, J. and Schulman, J. and Sutskever, I. and Cobbe, K.},
  journal={arXiv preprint arXiv:2305.20050},
  year={2023},
  doi={10.48550/arXiv.2305.20050}
}

@article{madaan2023self,
  title={Self-refine: Iterative refinement with self-feedback},
  author={Madaan, A. and Tandon, N. and Gupta, P. and Hallinan, S. and Gao, L. and Wiegreffe, S. and Alon, U. and Dziri, N. and Prabhumoye, S. and Yang, Y. and Gupta, S. and Majumder, B. P. and Hermann, K. and Welleck, S. and Yazdanbakhsh, A. and Clark, P.},
  journal={arXiv preprint arXiv:2303.17651},
  year={2023},
  doi={10.48550/arXiv.2303.17651}
}

@article{muennighoff2025s1,
  title={S1: Simple test-time scaling},
  author={Muennighoff, N. and Yang, Z. and Shi, W. and Li, X. L. and Fei-Fei, L. and Hajishirzi, H. and Zettlemoyer, L. and Liang, P. and Cand{\`e}s, E. and Hashimoto, T.},
  journal={arXiv preprint arXiv:2501.19393},
  year={2025},
  doi={10.48550/arXiv.2501.19393}
}

@misc{ng2025whats,
  title={What's next for {AI} agentic workflows},
  author={Ng, A.},
  howpublished={Video recording},
  year={2025},
  note={Retrieved January 24, 2025, from https://www.youtube.com/watch?v=sal78ACtGTc}
}

@misc{openai2024learning,
  title={Learning to reason with {LLMs}},
  author={{OpenAI Research}},
  year={2024},
  month={September},
  howpublished={OpenAI},
  note={Retrieved from https://openai.com/index/learning-to-reason-with-llms/}
}

@article{qin2024o1,
  title={O1 replication journey: A strategic progress report -- part 1},
  author={Qin, Y. and Li, X. and Zou, H. and Liu, Y. and Xia, S. and Huang, Z. and Ye, Y. and Yuan, W. and Liu, H. and Li, Y. and Liu, P.},
  journal={arXiv preprint arXiv:2410.18982},
  year={2024},
  doi={10.48550/arXiv.2410.18982}
}

@techreport{salahi2024more,
  title={More Effectively Searching Trees of Thought for Increased Reasoning Ability in Large Language Models},
  author={Salahi, K. and Gurusankar, P. and Edamadaka, S.},
  institution={CS224N},
  year={2024},
  url={https://web.stanford.edu/class/archive/cs/cs224n/cs224n.1244/final-projects/KamyarJohnSalahiPranavGurusankarSathyaEdamadaka.pdf}
}

@article{sami2024system,
  title={System for systematic literature review using multiple AI agents: Concept and an empirical evaluation},
  author={Sami, A. M. and Rasheed, Z. and Kemell, K.-K. and Waseem, M. and Kilamo, T. and Saari, M. and Duc, A. N. and Syst{\"a}, K. and Abrahamsson, P.},
  journal={arXiv preprint arXiv:2403.08399},
  year={2024},
  doi={10.48550/arXiv.2403.08399}
}

@article{shazeer2017outrageously,
  title={Outrageously large neural networks: The sparsely-gated mixture-of-experts layer},
  author={Shazeer, N. and Mirhoseini, A. and Maziarz, K. and Davis, A. and Le, Q. and Hinton, G. and Dean, J.},
  journal={arXiv preprint arXiv:1701.06538},
  year={2017},
  doi={10.48550/arXiv.1701.06538}
}

@article{stroebl2024inference,
  title={Inference scaling flaws: The limits of llm resampling with imperfect verifiers},
  author={Stroebl, B. and Kapoor, S. and Narayanan, A.},
  journal={arXiv preprint arXiv:2411.17501},
  year={2024},
  doi={10.48550/arXiv.2411.17501}
}

@article{valmeekam2024llms,
  title={LLMs still can't plan; can LRMs? A preliminary evaluation of openai's o1 on planbench},
  author={Valmeekam, K. and Stechly, K. and Kambhampati, S.},
  journal={arXiv preprint arXiv:2409.13373},
  year={2024},
  doi={10.48550/arXiv.2409.13373}
}

@article{wei2023chain,
  title={Chain-of-thought prompting elicits reasoning in large language models},
  author={Wei, J. and Wang, X. and Schuurmans, D. and Bosma, M. and Ichter, B. and Xia, F. and Chi, E. and Le, Q. and Zhou, D.},
  journal={arXiv preprint arXiv:2201.11903},
  year={2023},
  doi={10.48550/arXiv.2201.11903}
}

@article{wu2023autogen,
  title={AutoGen: Enabling next-gen LLM applications via multi-agent conversation},
  author={Wu, Q. and Bansal, G. and Zhang, J. and Wu, Y. and Li, B. and Zhu, E. and Jiang, L. and Zhang, X. and Zhang, S. and Liu, J. and Awadallah, A. H. and White, R. W. and Burger, D. and Wang, C.},
  journal={arXiv preprint arXiv:2308.08155},
  year={2023},
  doi={10.48550/arXiv.2308.08155}
}

@article{xu2024hallucination,
  title={Hallucination is inevitable: An innate limitation of large language models},
  author={Xu, Z. and Jain, S. and Kankanhalli, M.},
  journal={arXiv preprint arXiv:2401.11817},
  year={2024},
  doi={10.48550/arXiv.2401.11817}
}

@article{ye2025sop,
  title={SOP-Agent: Empower general purpose AI agent with domain-specific SOPs},
  author={Ye, A. and Ma, Q. and Chen, J. and Li, M. and Li, T. and Liu, F. and Mai, S. and Lu, M. and Bao, H. and You, Y.},
  journal={arXiv preprint arXiv:2501.09316},
  year={2025},
  doi={10.48550/arXiv.2501.09316}
}

@misc{openai2024gpt4o,
  title={GPT-4o},
  author={OpenAI},
  year={2024},
  note={\url{https://openai.com/index/gpt-4o-system-card/}}
}

@misc{openai2024gpt4omini,
  title={GPT-4o Mini},
  author={OpenAI},
  year={2024},
  note={\url{https://openai.com/index/gpt-4o-mini-advancing-cost-efficient-intelligence/}}
}

@misc{openai2024o3mini,
  title={o3-mini},
  author={OpenAI},
  year={2024},
  note={\url{https://openai.com/index/openai-o3-mini/}}
}

@misc{perplexity2025deepresearch,
  title={Introducing Perplexity Deep Research},
  author={Perplexity AI},
  year={2025},
  note={\url{https://www.perplexity.ai/hub/blog/introducing-perplexity-deep-research}}
}

@misc{openai2025deepresearch,
  title={OpenAI Deep Research},
  author={OpenAI},
  year={2025},
  note={\url{https://openai.com/index/introducing-deep-research/}}
}

@misc{gupta2025inkwell,
  title={InkwellAI: Hierarchical Multi-Agent System for Academic Writing},
  author={Asanshay Gupta},
  year={2025},
  note={\url{https://inkwellai.streamlit.app}}
}

@misc{choi2024picleelicitingdiversebehaviors,
      title={PICLe: Eliciting Diverse Behaviors from Large Language Models with Persona In-Context Learning}, 
      author={Hyeong Kyu Choi and Yixuan Li},
      year={2024},
      eprint={2405.02501},
      archivePrefix={arXiv},
      primaryClass={cs.CL},
      url={https://arxiv.org/abs/2405.02501}, 
}

@misc{li_towards_2024,
	title = {Towards goal-oriented prompt engineering for large language models: a survey},
	shorttitle = {Towards goal-oriented prompt engineering for large language models},
	url = {http://arxiv.org/abs/2401.14043},
	doi = {10.48550/arXiv.2401.14043},
	abstract = {Large Language Models (LLMs) have shown prominent performance in various downstream tasks and prompt engineering plays a pivotal role in optimizing LLMs' performance. This paper, not only as an overview of current prompt engineering methods, but also aims to highlight the limitation of designing prompts based on an anthropomorphic assumption that expects LLMs to think like humans. From our review of 50 representative studies, we demonstrate that a goal-oriented prompt formulation, which guides LLMs to follow established human logical thinking, significantly improves the performance of LLMs. Furthermore, We introduce a novel taxonomy that categorizes goal-oriented prompting methods into five interconnected stages and we demonstrate the broad applicability of our framework. With four future directions proposed, we hope to further emphasize the power and potential of goal-oriented prompt engineering in all fields.},
	urldate = {2025-01-25},
	publisher = {arXiv},
	author = {Li, Haochen and Leung, Jonathan and Shen, Zhiqi},
	month = sep,
	year = {2024},
	note = {arXiv:2401.14043},
	keywords = {Computer Science - Computation and Language, Computer Science - Artificial Intelligence},
}

@misc{stroebl_inference_2024,
	title = {Inference scaling flaws: the limits of llm resampling with imperfect verifiers},
	shorttitle = {Inference {Scaling} {fLaws}},
	url = {http://arxiv.org/abs/2411.17501},
	doi = {10.48550/arXiv.2411.17501},
	abstract = {Recent research has generated hope that inference scaling could allow weaker language models to match or exceed the accuracy of stronger models, such as by repeatedly sampling solutions to a coding problem until it passes unit tests. The central thesis of this paper is that there is no free lunch for inference scaling: indefinite accuracy improvement through resampling can only be realized if the "verifier" (in this case, a set of unit tests) is perfect. When the verifier is imperfect, as it almost always is in domains such as reasoning or coding (for example, unit tests have imperfect coverage), there is a nonzero probability of false positives: incorrect solutions that pass the verifier. Resampling cannot decrease this probability, so it imposes an upper bound to the accuracy of resampling-based inference scaling even with an infinite compute budget. We find that there is a very strong correlation between the model's single-sample accuracy (i.e. accuracy without unit tests) and its false positive rate on coding benchmarks HumanEval and MBPP, whose unit tests have limited coverage. Therefore, no amount of inference scaling of weaker models can enable them to match the single-sample accuracy of a sufficiently strong model (Fig. 1a). When we consider that false positives have a negative utility compared to abstaining from producing a solution, it bends the inference scaling curve further downward. Empirically, we find that the optimal number of samples can be less than 10 under realistic assumptions (Fig. 1b). Finally, we show that beyond accuracy, false positives may have other undesirable qualities, such as poor adherence to coding style conventions.},
	urldate = {2025-01-25},
	publisher = {arXiv},
	author = {Stroebl, Benedikt and Kapoor, Sayash and Narayanan, Arvind},
	month = dec,
	year = {2024},
	note = {arXiv:2411.17501},
	keywords = {Computer Science - Machine Learning, Computer Science - Artificial Intelligence},
}

@misc{madaan_self-refine:_2023,
	title = {Self-refine: iterative refinement with self-feedback},
	shorttitle = {Self-refine},
	url = {http://arxiv.org/abs/2303.17651},
	doi = {10.48550/arXiv.2303.17651},
	abstract = {Like humans, large language models (LLMs) do not always generate the best output on their first try. Motivated by how humans refine their written text, we introduce Self-Refine, an approach for improving initial outputs from LLMs through iterative feedback and refinement. The main idea is to generate an initial output using an LLMs; then, the same LLMs provides feedback for its output and uses it to refine itself, iteratively. Self-Refine does not require any supervised training data, additional training, or reinforcement learning, and instead uses a single LLM as the generator, refiner, and feedback provider. We evaluate Self-Refine across 7 diverse tasks, ranging from dialog response generation to mathematical reasoning, using state-of-the-art (GPT-3.5, ChatGPT, and GPT-4) LLMs. Across all evaluated tasks, outputs generated with Self-Refine are preferred by humans and automatic metrics over those generated with the same LLM using conventional one-step generation, improving by {\textasciitilde}20\% absolute on average in task performance. Our work demonstrates that even state-of-the-art LLMs like GPT-4 can be further improved at test time using our simple, standalone approach.},
	urldate = {2025-01-25},
	publisher = {arXiv},
	author = {Madaan, Aman and Tandon, Niket and Gupta, Prakhar and Hallinan, Skyler and Gao, Luyu and Wiegreffe, Sarah and Alon, Uri and Dziri, Nouha and Prabhumoye, Shrimai and Yang, Yiming and Gupta, Shashank and Majumder, Bodhisattwa Prasad and Hermann, Katherine and Welleck, Sean and Yazdanbakhsh, Amir and Clark, Peter},
	month = may,
	year = {2023},
	note = {arXiv:2303.17651},
	keywords = {Computer Science - Computation and Language, Computer Science - Artificial Intelligence, Computer Science - Machine Learning},
}

@misc{wu_autogen:_2023,
	title = {Autogen: enabling next-gen llm applications via multi-agent conversation},
	shorttitle = {Autogen},
	url = {http://arxiv.org/abs/2308.08155},
	doi = {10.48550/arXiv.2308.08155},
	abstract = {AutoGen is an open-source framework that allows developers to build LLM applications via multiple agents that can converse with each other to accomplish tasks. AutoGen agents are customizable, conversable, and can operate in various modes that employ combinations of LLMs, human inputs, and tools. Using AutoGen, developers can also flexibly define agent interaction behaviors. Both natural language and computer code can be used to program flexible conversation patterns for different applications. AutoGen serves as a generic infrastructure to build diverse applications of various complexities and LLM capacities. Empirical studies demonstrate the effectiveness of the framework in many example applications, with domains ranging from mathematics, coding, question answering, operations research, online decision-making, entertainment, etc.},
	urldate = {2025-01-25},
	publisher = {arXiv},
	author = {Wu, Qingyun and Bansal, Gagan and Zhang, Jieyu and Wu, Yiran and Li, Beibin and Zhu, Erkang and Jiang, Li and Zhang, Xiaoyun and Zhang, Shaokun and Liu, Jiale and Awadallah, Ahmed Hassan and White, Ryen W. and Burger, Doug and Wang, Chi},
	month = oct,
	year = {2023},
	note = {arXiv:2308.08155},
	keywords = {Computer Science - Artificial Intelligence, Computer Science - Computation and Language},
}

@misc{ye_sop-agent:_2025,
	title = {Sop-agent: empower general purpose ai agent with domain-specific sops},
	shorttitle = {Sop-agent},
	url = {http://arxiv.org/abs/2501.09316},
	doi = {10.48550/arXiv.2501.09316},
	abstract = {Despite significant advancements in general-purpose AI agents, several challenges still hinder their practical application in real-world scenarios. First, the limited planning capabilities of Large Language Models (LLM) restrict AI agents from effectively solving complex tasks that require long-horizon planning. Second, general-purpose AI agents struggle to efficiently utilize domain-specific knowledge and human expertise. In this paper, we introduce the Standard Operational Procedure-guided Agent (SOP-agent), a novel framework for constructing domain-specific agents through pseudocode-style Standard Operational Procedures (SOPs) written in natural language. Formally, we represent a SOP as a decision graph, which is traversed to guide the agent in completing tasks specified by the SOP. We conduct extensive experiments across tasks in multiple domains, including decision-making, search and reasoning, code generation, data cleaning, and grounded customer service. The SOP-agent demonstrates excellent versatility, achieving performance superior to general-purpose agent frameworks and comparable to domain-specific agent systems. Additionally, we introduce the Grounded Customer Service Benchmark, the first benchmark designed to evaluate the grounded decision-making capabilities of AI agents in customer service scenarios based on SOPs.},
	urldate = {2025-01-25},
	publisher = {arXiv},
	author = {Ye, Anbang and Ma, Qianran and Chen, Jia and Li, Muqi and Li, Tong and Liu, Fujiao and Mai, Siqi and Lu, Meichen and Bao, Haitao and You, Yang},
	month = jan,
	year = {2025},
	note = {arXiv:2501.09316},
	keywords = {Computer Science - Artificial Intelligence},
}

@misc{cheng_exploring_2024,
	title = {Exploring large language model based intelligent agents: definitions, methods, and prospects},
	shorttitle = {Exploring large language model based intelligent agents},
	url = {http://arxiv.org/abs/2401.03428},
	doi = {10.48550/arXiv.2401.03428},
	abstract = {Intelligent agents stand out as a potential path toward artificial general intelligence (AGI). Thus, researchers have dedicated significant effort to diverse implementations for them. Benefiting from recent progress in large language models (LLMs), LLM-based agents that use universal natural language as an interface exhibit robust generalization capabilities across various applications -- from serving as autonomous general-purpose task assistants to applications in coding, social, and economic domains, LLM-based agents offer extensive exploration opportunities. This paper surveys current research to provide an in-depth overview of LLM-based intelligent agents within single-agent and multi-agent systems. It covers their definitions, research frameworks, and foundational components such as their composition, cognitive and planning methods, tool utilization, and responses to environmental feedback. We also delve into the mechanisms of deploying LLM-based agents in multi-agent systems, including multi-role collaboration, message passing, and strategies to alleviate communication issues between agents. The discussions also shed light on popular datasets and application scenarios. We conclude by envisioning prospects for LLM-based agents, considering the evolving landscape of AI and natural language processing.},
	urldate = {2025-01-25},
	publisher = {arXiv},
	author = {Cheng, Yuheng and Zhang, Ceyao and Zhang, Zhengwen and Meng, Xiangrui and Hong, Sirui and Li, Wenhao and Wang, Zihao and Wang, Zekai and Yin, Feng and Zhao, Junhua and He, Xiuqiang},
	month = jan,
	year = {2024},
	note = {arXiv:2401.03428},
	keywords = {Computer Science - Artificial Intelligence, Computer Science - Multiagent Systems},
}

@misc{qin_o1_2024,
	title = {O1 replication journey: a strategic progress report -- part 1},
	shorttitle = {O1 replication journey},
	url = {http://arxiv.org/abs/2410.18982},
	doi = {10.48550/arXiv.2410.18982},
	abstract = {This paper introduces a pioneering approach to artificial intelligence research, embodied in our O1 Replication Journey. In response to the announcement of OpenAI's groundbreaking O1 model, we embark on a transparent, real-time exploration to replicate its capabilities while reimagining the process of conducting and communicating AI research. Our methodology addresses critical challenges in modern AI research, including the insularity of prolonged team-based projects, delayed information sharing, and the lack of recognition for diverse contributions. By providing comprehensive, real-time documentation of our replication efforts, including both successes and failures, we aim to foster open science, accelerate collective advancement, and lay the groundwork for AI-driven scientific discovery. Our research progress report diverges significantly from traditional research papers, offering continuous updates, full process transparency, and active community engagement throughout the research journey. Technologically, we proposed the journey learning paradigm, which encourages models to learn not just shortcuts, but the complete exploration process, including trial and error, reflection, and backtracking. With only 327 training samples and without any additional tricks, journey learning outperformed conventional supervised learning by over 8{\textbackslash}\% on the MATH dataset, demonstrating its extremely powerful potential. We believe this to be the most crucial component of O1 technology that we have successfully decoded. We share valuable resources including technical hypotheses and insights, cognitive exploration maps, custom-developed tools, etc at https://github.com/GAIR-NLP/O1-Journey.},
	urldate = {2025-01-25},
	publisher = {arXiv},
	author = {Qin, Yiwei and Li, Xuefeng and Zou, Haoyang and Liu, Yixiu and Xia, Shijie and Huang, Zhen and Ye, Yixin and Yuan, Weizhe and Liu, Hector and Li, Yuanzhi and Liu, Pengfei},
	month = oct,
	year = {2024},
	note = {arXiv:2410.18982},
	keywords = {Computer Science - Artificial Intelligence, Computer Science - Computation and Language},
}

@misc{valmeekam_llms_2024,
	title = {Llms still can't plan; can lrms? {A} preliminary evaluation of openai's o1 on planbench},
	shorttitle = {Llms still can't plan; can lrms?},
	url = {http://arxiv.org/abs/2409.13373},
	doi = {10.48550/arXiv.2409.13373},
	abstract = {The ability to plan a course of action that achieves a desired state of affairs has long been considered a core competence of intelligent agents and has been an integral part of AI research since its inception. With the advent of large language models (LLMs), there has been considerable interest in the question of whether or not they possess such planning abilities. PlanBench, an extensible benchmark we developed in 2022, soon after the release of GPT3, has remained an important tool for evaluating the planning abilities of LLMs. Despite the slew of new private and open source LLMs since GPT3, progress on this benchmark has been surprisingly slow. OpenAI claims that their recent o1 (Strawberry) model has been specifically constructed and trained to escape the normal limitations of autoregressive LLMs--making it a new kind of model: a Large Reasoning Model (LRM). Using this development as a catalyst, this paper takes a comprehensive look at how well current LLMs and new LRMs do on PlanBench. As we shall see, while o1's performance is a quantum improvement on the benchmark, outpacing the competition, it is still far from saturating it. This improvement also brings to the fore questions about accuracy, efficiency, and guarantees which must be considered before deploying such systems.},
	urldate = {2025-01-25},
	publisher = {arXiv},
	author = {Valmeekam, Karthik and Stechly, Kaya and Kambhampati, Subbarao},
	month = sep,
	year = {2024},
	note = {arXiv:2409.13373},
	keywords = {Computer Science - Artificial Intelligence, Computer Science - Computation and Language},
}

@misc{sami_system_2024,
	title = {System for systematic literature review using multiple {AI} agents: {Concept} and an empirical evaluation},
	shorttitle = {System for systematic literature review using multiple {AI} agents},
	url = {http://arxiv.org/abs/2403.08399},
	doi = {10.48550/arXiv.2403.08399},
	abstract = {Systematic Literature Reviews (SLRs) have become the foundation of evidence-based studies, enabling researchers to identify, classify, and combine existing studies based on specific research questions. Conducting an SLR is largely a manual process. Over the previous years, researchers have made significant progress in automating certain phases of the SLR process, aiming to reduce the effort and time needed to carry out high-quality SLRs. However, there is still a lack of AI agent-based models that automate the entire SLR process. To this end, we introduce a novel multi-AI agent model designed to fully automate the process of conducting an SLR. By utilizing the capabilities of Large Language Models (LLMs), our proposed model streamlines the review process, enhancing efficiency and accuracy. The model operates through a user-friendly interface where researchers input their topic, and in response, the model generates a search string used to retrieve relevant academic papers. Subsequently, an inclusive and exclusive filtering process is applied, focusing on titles relevant to the specific research area. The model then autonomously summarizes the abstracts of these papers, retaining only those directly related to the field of study. In the final phase, the model conducts a thorough analysis of the selected papers concerning predefined research questions. We also evaluated the proposed model by sharing it with ten competent software engineering researchers for testing and analysis. The researchers expressed strong satisfaction with the proposed model and provided feedback for further improvement. The code for this project can be found on the GitHub repository at https://github.com/GPT-Laboratory/SLR-automation.},
	urldate = {2025-01-25},
	publisher = {arXiv},
	author = {Sami, Abdul Malik and Rasheed, Zeeshan and Kemell, Kai-Kristian and Waseem, Muhammad and Kilamo, Terhi and Saari, Mika and Duc, Anh Nguyen and Systä, Kari and Abrahamsson, Pekka},
	month = mar,
	year = {2024},
	note = {arXiv:2403.08399},
	keywords = {Computer Science - Software Engineering},
}

@misc{besta_graph_2024,
	title = {Graph of thoughts: solving elaborate problems with large language models},
	shorttitle = {Graph of thoughts},
	url = {http://arxiv.org/abs/2308.09687},
	doi = {10.48550/arXiv.2308.09687},
	abstract = {We introduce Graph of Thoughts (GoT): a framework that advances prompting capabilities in large language models (LLMs) beyond those offered by paradigms such as Chain-of-Thought or Tree of Thoughts (ToT). The key idea and primary advantage of GoT is the ability to model the information generated by an LLM as an arbitrary graph, where units of information ("LLM thoughts") are vertices, and edges correspond to dependencies between these vertices. This approach enables combining arbitrary LLM thoughts into synergistic outcomes, distilling the essence of whole networks of thoughts, or enhancing thoughts using feedback loops. We illustrate that GoT offers advantages over state of the art on different tasks, for example increasing the quality of sorting by 62\% over ToT, while simultaneously reducing costs by {\textgreater}31\%. We ensure that GoT is extensible with new thought transformations and thus can be used to spearhead new prompting schemes. This work brings the LLM reasoning closer to human thinking or brain mechanisms such as recurrence, both of which form complex networks.},
	urldate = {2025-01-25},
	publisher = {arXiv},
	author = {Besta, Maciej and Blach, Nils and Kubicek, Ales and Gerstenberger, Robert and Podstawski, Michal and Gianinazzi, Lukas and Gajda, Joanna and Lehmann, Tomasz and Niewiadomski, Hubert and Nyczyk, Piotr and Hoefler, Torsten},
	month = feb,
	year = {2024},
	note = {arXiv:2308.09687},
	keywords = {Computer Science - Computation and Language, Computer Science - Artificial Intelligence, Computer Science - Machine Learning},
}

@misc{brahman_art_2024,
	title = {The art of saying no: contextual noncompliance in language models},
	shorttitle = {The art of saying no},
	url = {http://arxiv.org/abs/2407.12043},
	doi = {10.48550/arXiv.2407.12043},
	abstract = {Chat-based language models are designed to be helpful, yet they should not comply with every user request. While most existing work primarily focuses on refusal of "unsafe" queries, we posit that the scope of noncompliance should be broadened. We introduce a comprehensive taxonomy of contextual noncompliance describing when and how models should not comply with user requests. Our taxonomy spans a wide range of categories including incomplete, unsupported, indeterminate, and humanizing requests (in addition to unsafe requests). To test noncompliance capabilities of language models, we use this taxonomy to develop a new evaluation suite of 1000 noncompliance prompts. We find that most existing models show significantly high compliance rates in certain previously understudied categories with models like GPT-4 incorrectly complying with as many as 30\% of requests. To address these gaps, we explore different training strategies using a synthetically-generated training set of requests and expected noncompliant responses. Our experiments demonstrate that while direct finetuning of instruction-tuned models can lead to both over-refusal and a decline in general capabilities, using parameter efficient methods like low rank adapters helps to strike a good balance between appropriate noncompliance and other capabilities.},
	urldate = {2025-01-25},
	publisher = {arXiv},
	author = {Brahman, Faeze and Kumar, Sachin and Balachandran, Vidhisha and Dasigi, Pradeep and Pyatkin, Valentina and Ravichander, Abhilasha and Wiegreffe, Sarah and Dziri, Nouha and Chandu, Khyathi and Hessel, Jack and Tsvetkov, Yulia and Smith, Noah A. and Choi, Yejin and Hajishirzi, Hannaneh},
	month = nov,
	year = {2024},
	note = {arXiv:2407.12043},
	keywords = {Computer Science - Computation and Language, Computer Science - Artificial Intelligence, Computer Science - Human-Computer Interaction},
}

@misc{noauthor_building_nodate,
	title = {Building effective agents},
	url = {https://www.anthropic.com/research/building-effective-agents},
	abstract = {A post for developers with advice and workflows for building effective AI agents},
	language = {en},
	urldate = {2025-01-25},
}

@misc{ng_whats_nodate,
	title = {What's next for {AI} agentic workflows},
	url = {https://www.youtube.com/watch?v=sal78ACtGTc},
	urldate = {2025-01-25},
	author = {Ng, Andrew},
}

@misc{lightman_lets_2023,
	title = {Let's verify step by step},
	url = {http://arxiv.org/abs/2305.20050},
	doi = {10.48550/arXiv.2305.20050},
	abstract = {In recent years, large language models have greatly improved in their ability to perform complex multi-step reasoning. However, even state-of-the-art models still regularly produce logical mistakes. To train more reliable models, we can turn either to outcome supervision, which provides feedback for a final result, or process supervision, which provides feedback for each intermediate reasoning step. Given the importance of training reliable models, and given the high cost of human feedback, it is important to carefully compare the both methods. Recent work has already begun this comparison, but many questions still remain. We conduct our own investigation, finding that process supervision significantly outperforms outcome supervision for training models to solve problems from the challenging MATH dataset. Our process-supervised model solves 78\% of problems from a representative subset of the MATH test set. Additionally, we show that active learning significantly improves the efficacy of process supervision. To support related research, we also release PRM800K, the complete dataset of 800,000 step-level human feedback labels used to train our best reward model.},
	urldate = {2025-02-07},
	publisher = {arXiv},
	author = {Lightman, Hunter and Kosaraju, Vineet and Burda, Yura and Edwards, Harri and Baker, Bowen and Lee, Teddy and Leike, Jan and Schulman, John and Sutskever, Ilya and Cobbe, Karl},
	month = may,
	year = {2023},
	note = {arXiv:2305.20050},
	keywords = {Computer Science - Machine Learning, Computer Science - Artificial Intelligence, Computer Science - Computation and Language},
}

@misc{wei_chain--thought_2023,
	title = {Chain-of-thought prompting elicits reasoning in large language models},
	url = {http://arxiv.org/abs/2201.11903},
	doi = {10.48550/arXiv.2201.11903},
	abstract = {We explore how generating a chain of thought -- a series of intermediate reasoning steps -- significantly improves the ability of large language models to perform complex reasoning. In particular, we show how such reasoning abilities emerge naturally in sufficiently large language models via a simple method called chain of thought prompting, where a few chain of thought demonstrations are provided as exemplars in prompting. Experiments on three large language models show that chain of thought prompting improves performance on a range of arithmetic, commonsense, and symbolic reasoning tasks. The empirical gains can be striking. For instance, prompting a 540B-parameter language model with just eight chain of thought exemplars achieves state of the art accuracy on the GSM8K benchmark of math word problems, surpassing even finetuned GPT-3 with a verifier.},
	urldate = {2025-02-07},
	publisher = {arXiv},
	author = {Wei, Jason and Wang, Xuezhi and Schuurmans, Dale and Bosma, Maarten and Ichter, Brian and Xia, Fei and Chi, Ed and Le, Quoc and Zhou, Denny},
	month = jan,
	year = {2023},
	note = {arXiv:2201.11903},
	keywords = {Computer Science - Computation and Language, Computer Science - Artificial Intelligence},
}

@misc{lewis_retrieval-augmented_2021,
	title = {Retrieval-augmented generation for knowledge-intensive nlp tasks},
	url = {http://arxiv.org/abs/2005.11401},
	doi = {10.48550/arXiv.2005.11401},
	abstract = {Large pre-trained language models have been shown to store factual knowledge in their parameters, and achieve state-of-the-art results when fine-tuned on downstream NLP tasks. However, their ability to access and precisely manipulate knowledge is still limited, and hence on knowledge-intensive tasks, their performance lags behind task-specific architectures. Additionally, providing provenance for their decisions and updating their world knowledge remain open research problems. Pre-trained models with a differentiable access mechanism to explicit non-parametric memory can overcome this issue, but have so far been only investigated for extractive downstream tasks. We explore a general-purpose fine-tuning recipe for retrieval-augmented generation (RAG) -- models which combine pre-trained parametric and non-parametric memory for language generation. We introduce RAG models where the parametric memory is a pre-trained seq2seq model and the non-parametric memory is a dense vector index of Wikipedia, accessed with a pre-trained neural retriever. We compare two RAG formulations, one which conditions on the same retrieved passages across the whole generated sequence, the other can use different passages per token. We fine-tune and evaluate our models on a wide range of knowledge-intensive NLP tasks and set the state-of-the-art on three open domain QA tasks, outperforming parametric seq2seq models and task-specific retrieve-and-extract architectures. For language generation tasks, we find that RAG models generate more specific, diverse and factual language than a state-of-the-art parametric-only seq2seq baseline.},
	urldate = {2025-02-07},
	publisher = {arXiv},
	author = {Lewis, Patrick and Perez, Ethan and Piktus, Aleksandra and Petroni, Fabio and Karpukhin, Vladimir and Goyal, Naman and Küttler, Heinrich and Lewis, Mike and Yih, Wen-tau and Rocktäschel, Tim and Riedel, Sebastian and Kiela, Douwe},
	month = apr,
	year = {2021},
	note = {arXiv:2005.11401},
	keywords = {Computer Science - Computation and Language, Computer Science - Machine Learning},
}

@misc{muennighoff_s1:_2025,
	title = {S1: simple test-time scaling},
	shorttitle = {S1},
	url = {http://arxiv.org/abs/2501.19393},
	doi = {10.48550/arXiv.2501.19393},
	abstract = {Test-time scaling is a promising new approach to language modeling that uses extra test-time compute to improve performance. Recently, OpenAI's o1 model showed this capability but did not publicly share its methodology, leading to many replication efforts. We seek the simplest approach to achieve test-time scaling and strong reasoning performance. First, we curate a small dataset s1K of 1,000 questions paired with reasoning traces relying on three criteria we validate through ablations: difficulty, diversity, and quality. Second, we develop budget forcing to control test-time compute by forcefully terminating the model's thinking process or lengthening it by appending "Wait" multiple times to the model's generation when it tries to end. This can lead the model to double-check its answer, often fixing incorrect reasoning steps. After supervised finetuning the Qwen2.5-32B-Instruct language model on s1K and equipping it with budget forcing, our model s1-32B exceeds o1-preview on competition math questions by up to 27\% (MATH and AIME24). Further, scaling s1-32B with budget forcing allows extrapolating beyond its performance without test-time intervention: from 50\% to 57\% on AIME24. Our model, data, and code are open-source at https://github.com/simplescaling/s1},
	urldate = {2025-02-07},
	publisher = {arXiv},
	author = {Muennighoff, Niklas and Yang, Zitong and Shi, Weijia and Li, Xiang Lisa and Fei-Fei, Li and Hajishirzi, Hannaneh and Zettlemoyer, Luke and Liang, Percy and Candès, Emmanuel and Hashimoto, Tatsunori},
	month = feb,
	year = {2025},
	note = {arXiv:2501.19393},
	keywords = {Computer Science - Computation and Language, Computer Science - Artificial Intelligence, Computer Science - Machine Learning},
}

@misc{havrilla_teaching_2024,
	title = {Teaching large language models to reason with reinforcement learning},
	url = {http://arxiv.org/abs/2403.04642},
	doi = {10.48550/arXiv.2403.04642},
	abstract = {Reinforcement Learning from Human Feedback ({\textbackslash}textbf\{RLHF\}) has emerged as a dominant approach for aligning LLM outputs with human preferences. Inspired by the success of RLHF, we study the performance of multiple algorithms that learn from feedback (Expert Iteration, Proximal Policy Optimization ({\textbackslash}textbf\{PPO\}), Return-Conditioned RL) on improving LLM reasoning capabilities. We investigate both sparse and dense rewards provided to the LLM both heuristically and via a learned reward model. We additionally start from multiple model sizes and initializations both with and without supervised fine-tuning ({\textbackslash}textbf\{SFT\}) data. Overall, we find all algorithms perform comparably, with Expert Iteration performing best in most cases. Surprisingly, we find the sample complexity of Expert Iteration is similar to that of PPO, requiring at most on the order of \$10{\textasciicircum}6\$ samples to converge from a pretrained checkpoint. We investigate why this is the case, concluding that during RL training models fail to explore significantly beyond solutions already produced by SFT models. Additionally, we discuss a trade off between maj@1 and pass@96 metric performance during SFT training and how conversely RL training improves both simultaneously. We then conclude by discussing the implications of our findings for RLHF and the future role of RL in LLM fine-tuning.},
	urldate = {2025-02-07},
	publisher = {arXiv},
	author = {Havrilla, Alex and Du, Yuqing and Raparthy, Sharath Chandra and Nalmpantis, Christoforos and Dwivedi-Yu, Jane and Zhuravinskyi, Maksym and Hambro, Eric and Sukhbaatar, Sainbayar and Raileanu, Roberta},
	month = mar,
	year = {2024},
	note = {arXiv:2403.04642},
	keywords = {Computer Science - Machine Learning},
}

@misc{openai_research_learning_2024,
	title = {Learning to reason with {LLMs}},
	url = {https://openai.com/index/learning-to-reason-with-llms/},
	journal = {OpenAI},
	author = {{OpenAI Research}},
	month = sep,
	year = {2024},
}

@inproceedings{salahi_more_2024,
	title = {More {Effectively} {Searching} {Trees} of {Thought} for {Increased} {Reasoning} {Ability} in {Large} {Language} {Models}},
	url = {https://web.stanford.edu/class/archive/cs/cs224n/cs224n.1244/final-projects/KamyarJohnSalahiPranavGurusankarSathyaEdamadaka.pdf},
	author = {Salahi, Kamyar and Gurusankar, Pranav and Edamadaka, Sathya},
	year = {2024},
}

@misc{jiang_mixtral_2024,
	title = {Mixtral of experts},
	url = {http://arxiv.org/abs/2401.04088},
	doi = {10.48550/arXiv.2401.04088},
	abstract = {We introduce Mixtral 8x7B, a Sparse Mixture of Experts (SMoE) language model. Mixtral has the same architecture as Mistral 7B, with the difference that each layer is composed of 8 feedforward blocks (i.e. experts). For every token, at each layer, a router network selects two experts to process the current state and combine their outputs. Even though each token only sees two experts, the selected experts can be different at each timestep. As a result, each token has access to 47B parameters, but only uses 13B active parameters during inference. Mixtral was trained with a context size of 32k tokens and it outperforms or matches Llama 2 70B and GPT-3.5 across all evaluated benchmarks. In particular, Mixtral vastly outperforms Llama 2 70B on mathematics, code generation, and multilingual benchmarks. We also provide a model fine-tuned to follow instructions, Mixtral 8x7B - Instruct, that surpasses GPT-3.5 Turbo, Claude-2.1, Gemini Pro, and Llama 2 70B - chat model on human benchmarks. Both the base and instruct models are released under the Apache 2.0 license.},
	urldate = {2025-02-07},
	publisher = {arXiv},
	author = {Jiang, Albert Q. and Sablayrolles, Alexandre and Roux, Antoine and Mensch, Arthur and Savary, Blanche and Bamford, Chris and Chaplot, Devendra Singh and Casas, Diego de las and Hanna, Emma Bou and Bressand, Florian and Lengyel, Gianna and Bour, Guillaume and Lample, Guillaume and Lavaud, Lélio Renard and Saulnier, Lucile and Lachaux, Marie-Anne and Stock, Pierre and Subramanian, Sandeep and Yang, Sophia and Antoniak, Szymon and Scao, Teven Le and Gervet, Théophile and Lavril, Thibaut and Wang, Thomas and Lacroix, Timothée and Sayed, William El},
	month = jan,
	year = {2024},
	note = {arXiv:2401.04088 
version: 1},
	keywords = {Computer Science - Machine Learning, Computer Science - Computation and Language},
}

@misc{shazeer_outrageously_2017,
	title = {Outrageously large neural networks: the sparsely-gated mixture-of-experts layer},
	shorttitle = {Outrageously large neural networks},
	url = {http://arxiv.org/abs/1701.06538},
	doi = {10.48550/arXiv.1701.06538},
	abstract = {The capacity of a neural network to absorb information is limited by its number of parameters. Conditional computation, where parts of the network are active on a per-example basis, has been proposed in theory as a way of dramatically increasing model capacity without a proportional increase in computation. In practice, however, there are significant algorithmic and performance challenges. In this work, we address these challenges and finally realize the promise of conditional computation, achieving greater than 1000x improvements in model capacity with only minor losses in computational efficiency on modern GPU clusters. We introduce a Sparsely-Gated Mixture-of-Experts layer (MoE), consisting of up to thousands of feed-forward sub-networks. A trainable gating network determines a sparse combination of these experts to use for each example. We apply the MoE to the tasks of language modeling and machine translation, where model capacity is critical for absorbing the vast quantities of knowledge available in the training corpora. We present model architectures in which a MoE with up to 137 billion parameters is applied convolutionally between stacked LSTM layers. On large language modeling and machine translation benchmarks, these models achieve significantly better results than state-of-the-art at lower computational cost.},
	urldate = {2025-02-07},
	publisher = {arXiv},
	author = {Shazeer, Noam and Mirhoseini, Azalia and Maziarz, Krzysztof and Davis, Andy and Le, Quoc and Hinton, Geoffrey and Dean, Jeff},
	month = jan,
	year = {2017},
	note = {arXiv:1701.06538},
	keywords = {Computer Science - Machine Learning, Computer Science - Computation and Language, Computer Science - Neural and Evolutionary Computing, Statistics - Machine Learning},
}

@misc{xu_hallucination_2024,
	title = {Hallucination is inevitable: an innate limitation of large language models},
	shorttitle = {Hallucination is {Inevitable}},
	url = {http://arxiv.org/abs/2401.11817},
	doi = {10.48550/arXiv.2401.11817},
	abstract = {Hallucination has been widely recognized to be a significant drawback for large language models (LLMs). There have been many works that attempt to reduce the extent of hallucination. These efforts have mostly been empirical so far, which cannot answer the fundamental question whether it can be completely eliminated. In this paper, we formalize the problem and show that it is impossible to eliminate hallucination in LLMs. Specifically, we define a formal world where hallucination is defined as inconsistencies between a computable LLM and a computable ground truth function. By employing results from learning theory, we show that LLMs cannot learn all of the computable functions and will therefore always hallucinate. Since the formal world is a part of the real world which is much more complicated, hallucinations are also inevitable for real world LLMs. Furthermore, for real world LLMs constrained by provable time complexity, we describe the hallucination-prone tasks and empirically validate our claims. Finally, using the formal world framework, we discuss the possible mechanisms and efficacies of existing hallucination mitigators as well as the practical implications on the safe deployment of LLMs.},
	urldate = {2025-02-07},
	publisher = {arXiv},
	author = {Xu, Ziwei and Jain, Sanjay and Kankanhalli, Mohan},
	month = jan,
	year = {2024},
	note = {arXiv:2401.11817},
	keywords = {Computer Science - Computation and Language, Computer Science - Artificial Intelligence, Computer Science - Machine Learning},
}

@misc{prasad_adapt:_2024,
	title = {Adapt: as-needed decomposition and planning with language models},
	shorttitle = {Adapt},
	url = {http://arxiv.org/abs/2311.05772},
	doi = {10.48550/arXiv.2311.05772},
	abstract = {Large Language Models (LLMs) are increasingly being used for interactive decision-making tasks requiring planning and adapting to the environment. Recent works employ LLMs-as-agents in broadly two ways: iteratively determining the next action (iterative executors) or generating plans and executing sub-tasks using LLMs (plan-and-execute). However, these methods struggle with task complexity, as the inability to execute any sub-task may lead to task failure. To address these shortcomings, we introduce As-Needed Decomposition and Planning for complex Tasks (ADaPT), an approach that explicitly plans and decomposes complex sub-tasks as-needed, i.e., when the LLM is unable to execute them. ADaPT recursively decomposes sub-tasks to adapt to both task complexity and LLM capability. Our results demonstrate that ADaPT substantially outperforms established strong baselines, achieving success rates up to 28.3\% higher in ALFWorld, 27\% in WebShop, and 33\% in TextCraft -- a novel compositional dataset that we introduce. Through extensive analysis, we illustrate the importance of multilevel decomposition and establish that ADaPT dynamically adjusts to the capabilities of the executor LLM as well as to task complexity.},
	urldate = {2025-03-15},
	publisher = {arXiv},
	author = {Prasad, Archiki and Koller, Alexander and Hartmann, Mareike and Clark, Peter and Sabharwal, Ashish and Bansal, Mohit and Khot, Tushar},
	month = apr,
	year = {2024},
	note = {arXiv:2311.05772},
	keywords = {Computer Science - Artificial Intelligence, Computer Science - Computation and Language, Computer Science - Machine Learning},
}

@misc{wataoka2024selfpreferencebiasllmasajudge,
      title={Self-Preference Bias in LLM-as-a-Judge}, 
      author={Koki Wataoka and Tsubasa Takahashi and Ryokan Ri},
      year={2024},
      eprint={2410.21819},
      archivePrefix={arXiv},
      primaryClass={cs.CL},
      url={https://arxiv.org/abs/2410.21819}, 
}


@misc{hsieh2024middlecalibratingpositionalattention,
      title={Found in the Middle: Calibrating Positional Attention Bias Improves Long Context Utilization}, 
      author={Cheng-Yu Hsieh and Yung-Sung Chuang and Chun-Liang Li and Zifeng Wang and Long T. Le and Abhishek Kumar and James Glass and Alexander Ratner and Chen-Yu Lee and Ranjay Krishna and Tomas Pfister},
      year={2024},
      eprint={2406.16008},
      archivePrefix={arXiv},
      primaryClass={cs.CL},
      url={https://arxiv.org/abs/2406.16008}, 
}

@misc{sun2024enhancingdocumentlevelargumentextraction,
      title={Enhancing Document-level Argument Extraction with Definition-augmented Heuristic-driven Prompting for LLMs}, 
      author={Tongyue Sun and Jiayi Xiao},
      year={2024},
      eprint={2409.00214},
      archivePrefix={arXiv},
      primaryClass={cs.CL},
      url={https://arxiv.org/abs/2409.00214}, 
}

@misc{chen2024understandingindividualagentimportance,
      title={Understanding Individual Agent Importance in Multi-Agent System via Counterfactual Reasoning}, 
      author={Jianming Chen and Yawen Wang and Junjie Wang and Xiaofei Xie and jun Hu and Qing Wang and Fanjiang Xu},
      year={2024},
      eprint={2412.15619},
      archivePrefix={arXiv},
      primaryClass={cs.AI},
      url={https://arxiv.org/abs/2412.15619}, 
}

@misc{chen2025repromptplanningautomaticprompt,
      title={RePrompt: Planning by Automatic Prompt Engineering for Large Language Models Agents}, 
      author={Weizhe Chen and Sven Koenig and Bistra Dilkina},
      year={2025},
      eprint={2406.11132},
      archivePrefix={arXiv},
      primaryClass={cs.CL},
      url={https://arxiv.org/abs/2406.11132}, 
}

@misc{wang2023promptagentstrategicplanninglanguage,
      title={PromptAgent: Strategic Planning with Language Models Enables Expert-level Prompt Optimization}, 
      author={Xinyuan Wang and Chenxi Li and Zhen Wang and Fan Bai and Haotian Luo and Jiayou Zhang and Nebojsa Jojic and Eric P. Xing and Zhiting Hu},
      year={2023},
      eprint={2310.16427},
      archivePrefix={arXiv},
      primaryClass={cs.CL},
      url={https://arxiv.org/abs/2310.16427}, 
}

@misc{fourney2024magenticonegeneralistmultiagentsolving,
      title={Magentic-One: A Generalist Multi-Agent System for Solving Complex Tasks}, 
      author={Adam Fourney and Gagan Bansal and Hussein Mozannar and Cheng Tan and Eduardo Salinas and Erkang and Zhu and Friederike Niedtner and Grace Proebsting and Griffin Bassman and Jack Gerrits and Jacob Alber and Peter Chang and Ricky Loynd and Robert West and Victor Dibia and Ahmed Awadallah and Ece Kamar and Rafah Hosn and Saleema Amershi},
      year={2024},
      eprint={2411.04468},
      archivePrefix={arXiv},
      primaryClass={cs.AI},
      url={https://arxiv.org/abs/2411.04468}, 
}

@misc{shah2024agents,
      title={Agents Are Not Enough}, 
      author={Chirag Shah and Ryen W. White},
      year={2024},
      eprint={2412.16241},
      archivePrefix={arXiv},
      primaryClass={cs.AI},
      url={https://arxiv.org/abs/2412.16241}, 
}

@misc{wei2022finetunedlanguagemodelszeroshot,
      title={Finetuned Language Models Are Zero-Shot Learners}, 
      author={Jason Wei and Maarten Bosma and Vincent Y. Zhao and Kelvin Guu and Adams Wei Yu and Brian Lester and Nan Du and Andrew M. Dai and Quoc V. Le},
      year={2022},
      eprint={2109.01652},
      archivePrefix={arXiv},
      primaryClass={cs.CL},
      url={https://arxiv.org/abs/2109.01652}, 
}

@misc{chung2022scalinginstructionfinetunedlanguagemodels,
      title={Scaling Instruction-Finetuned Language Models}, 
      author={Hyung Won Chung and Le Hou and Shayne Longpre and Barret Zoph and Yi Tay and William Fedus and Yunxuan Li and Xuezhi Wang and Mostafa Dehghani and Siddhartha Brahma and Albert Webson and Shixiang Shane Gu and Zhuyun Dai and Mirac Suzgun and Xinyun Chen and Aakanksha Chowdhery and Alex Castro-Ros and Marie Pellat and Kevin Robinson and Dasha Valter and Sharan Narang and Gaurav Mishra and Adams Yu and Vincent Zhao and Yanping Huang and Andrew Dai and Hongkun Yu and Slav Petrov and Ed H. Chi and Jeff Dean and Jacob Devlin and Adam Roberts and Denny Zhou and Quoc V. Le and Jason Wei},
      year={2022},
      eprint={2210.11416},
      archivePrefix={arXiv},
      primaryClass={cs.LG},
      url={https://arxiv.org/abs/2210.11416}, 
}

@misc{chu2024professionalagentsevolving,
      title={Professional Agents -- Evolving Large Language Models into Autonomous Experts with Human-Level Competencies}, 
      author={Zhixuan Chu and Yan Wang and Feng Zhu and Lu Yu and Longfei Li and Jinjie Gu},
      year={2024},
      eprint={2402.03628},
      archivePrefix={arXiv},
      primaryClass={cs.CL},
      url={https://arxiv.org/abs/2402.03628}, 
}

@misc{zhang2022automaticchainthoughtprompting,
      title={Automatic Chain of Thought Prompting in Large Language Models}, 
      author={Zhuosheng Zhang and Aston Zhang and Mu Li and Alex Smola},
      year={2022},
      eprint={2210.03493},
      archivePrefix={arXiv},
      primaryClass={cs.CL},
      url={https://arxiv.org/abs/2210.03493}, 
}

@misc{liang2023holisticevaluationlanguagemodels,
      title={Holistic Evaluation of Language Models}, 
      author={Percy Liang and Rishi Bommasani and Tony Lee and Dimitris Tsipras and Dilara Soylu and Michihiro Yasunaga and Yian Zhang and Deepak Narayanan and Yuhuai Wu and Ananya Kumar and Benjamin Newman and Binhang Yuan and Bobby Yan and Ce Zhang and Christian Cosgrove and Christopher D. Manning and Christopher Ré and Diana Acosta-Navas and Drew A. Hudson and Eric Zelikman and Esin Durmus and Faisal Ladhak and Frieda Rong and Hongyu Ren and Huaxiu Yao and Jue Wang and Keshav Santhanam and Laurel Orr and Lucia Zheng and Mert Yuksekgonul and Mirac Suzgun and Nathan Kim and Neel Guha and Niladri Chatterji and Omar Khattab and Peter Henderson and Qian Huang and Ryan Chi and Sang Michael Xie and Shibani Santurkar and Surya Ganguli and Tatsunori Hashimoto and Thomas Icard and Tianyi Zhang and Vishrav Chaudhary and William Wang and Xuechen Li and Yifan Mai and Yuhui Zhang and Yuta Koreeda},
      year={2023},
      eprint={2211.09110},
      archivePrefix={arXiv},
      primaryClass={cs.CL},
      url={https://arxiv.org/abs/2211.09110}, 
}

@misc{jimenez2024swebenchlanguagemodelsresolve,
      title={SWE-bench: Can Language Models Resolve Real-World GitHub Issues?}, 
      author={Carlos E. Jimenez and John Yang and Alexander Wettig and Shunyu Yao and Kexin Pei and Ofir Press and Karthik Narasimhan},
      year={2024},
      eprint={2310.06770},
      archivePrefix={arXiv},
      primaryClass={cs.CL},
      url={https://arxiv.org/abs/2310.06770}, 
}

@misc{jimenez2024swebenchlanguagemodelsresolve,
      title={SWE-bench: Can Language Models Resolve Real-World GitHub Issues?}, 
      author={Carlos E. Jimenez and John Yang and Alexander Wettig and Shunyu Yao and Kexin Pei and Ofir Press and Karthik Narasimhan},
      year={2024},
      eprint={2310.06770},
      archivePrefix={arXiv},
      primaryClass={cs.CL},
      url={https://arxiv.org/abs/2310.06770}, 
}

@misc{
wang2024gta,
title={{GTA}: A Benchmark for General Tool Agents},
author={Jize Wang and Ma Zerun and Yining Li and Songyang Zhang and Cailian Chen and Kai Chen and Xinyi Le},
booktitle={The Thirty-eight Conference on Neural Information Processing Systems Datasets and Benchmarks Track},
year={2024},
url={https://openreview.net/forum?id=akEt8QAa6V}
}

@misc{gao2024llmbasednlgevaluationcurrent,
      title={LLM-based NLG Evaluation: Current Status and Challenges}, 
      author={Mingqi Gao and Xinyu Hu and Jie Ruan and Xiao Pu and Xiaojun Wan},
      year={2024},
      eprint={2402.01383},
      archivePrefix={arXiv},
      primaryClass={cs.CL},
      url={https://arxiv.org/abs/2402.01383}, 
}

@misc{shankar2024validatesvalidatorsaligningllmassisted,
      title={Who Validates the Validators? Aligning LLM-Assisted Evaluation of LLM Outputs with Human Preferences}, 
      author={Shreya Shankar and J. D. Zamfirescu-Pereira and Björn Hartmann and Aditya G. Parameswaran and Ian Arawjo},
      year={2024},
      eprint={2404.12272},
      archivePrefix={arXiv},
      primaryClass={cs.HC},
      url={https://arxiv.org/abs/2404.12272}, 
}

@article{Nasution2024ChatGPTLC,
  title={ChatGPT Label: Comparing the Quality of Human-Generated and LLM-Generated Annotations in Low-Resource Language NLP Tasks},
  author={Arbi Haza Nasution and Aytuğ Onan},
  journal={IEEE Access},
  year={2024},
  volume={12},
  pages={71876-71900},
  url={https://api.semanticscholar.org/CorpusID:269942327}
}

@article{Lin2004ROUGEAP,
  title={ROUGE: A Package for Automatic Evaluation of Summaries},
  author={Chin-Yew Lin},
  booktitle={Annual Meeting of the Association for Computational Linguistics},
  year={2004},
  url={https://api.semanticscholar.org/CorpusID:964287}
}

@article{Papineni2002BleuAM,
  title={Bleu: a Method for Automatic Evaluation of Machine Translation},
  author={Kishore Papineni and Salim Roukos and Todd Ward and Wei-Jing Zhu},
  booktitle={Annual Meeting of the Association for Computational Linguistics},
  year={2002},
  url={https://api.semanticscholar.org/CorpusID:11080756}
}

@misc{chiang2023largelanguagemodelsalternative,
      title={Can Large Language Models Be an Alternative to Human Evaluations?}, 
      author={Cheng-Han Chiang and Hung-yi Lee},
      year={2023},
      eprint={2305.01937},
      archivePrefix={arXiv},
      primaryClass={cs.CL},
      url={https://arxiv.org/abs/2305.01937}, 
}

@misc{li2025generationjudgmentopportunitieschallenges,
      title={From Generation to Judgment: Opportunities and Challenges of LLM-as-a-judge}, 
      author={Dawei Li and Bohan Jiang and Liangjie Huang and Alimohammad Beigi and Chengshuai Zhao and Zhen Tan and Amrita Bhattacharjee and Yuxuan Jiang and Canyu Chen and Tianhao Wu and Kai Shu and Lu Cheng and Huan Liu},
      year={2025},
      eprint={2411.16594},
      archivePrefix={arXiv},
      primaryClass={cs.AI},
      url={https://arxiv.org/abs/2411.16594}, 
}

@misc{schroeder2025trustllmjudgmentsreliability,
      title={Can You Trust LLM Judgments? Reliability of LLM-as-a-Judge}, 
      author={Kayla Schroeder and Zach Wood-Doughty},
      year={2025},
      eprint={2412.12509},
      archivePrefix={arXiv},
      primaryClass={cs.CL},
      url={https://arxiv.org/abs/2412.12509}, 
}

@misc{ye2024justiceprejudicequantifyingbiases,
      title={Justice or Prejudice? Quantifying Biases in LLM-as-a-Judge}, 
      author={Jiayi Ye and Yanbo Wang and Yue Huang and Dongping Chen and Qihui Zhang and Nuno Moniz and Tian Gao and Werner Geyer and Chao Huang and Pin-Yu Chen and Nitesh V Chawla and Xiangliang Zhang},
      year={2024},
      eprint={2410.02736},
      archivePrefix={arXiv},
      primaryClass={cs.CL},
      url={https://arxiv.org/abs/2410.02736}, 
}

@misc{dong2024llmpersonalizedjudge,
      title={Can LLM be a Personalized Judge?}, 
      author={Yijiang River Dong and Tiancheng Hu and Nigel Collier},
      year={2024},
      eprint={2406.11657},
      archivePrefix={arXiv},
      primaryClass={cs.CL},
      url={https://arxiv.org/abs/2406.11657}, 
}

@misc{snell2024scalingllmtesttimecompute,
      title={Scaling LLM Test-Time Compute Optimally can be More Effective than Scaling Model Parameters}, 
      author={Charlie Snell and Jaehoon Lee and Kelvin Xu and Aviral Kumar},
      year={2024},
      eprint={2408.03314},
      archivePrefix={arXiv},
      primaryClass={cs.LG},
      url={https://arxiv.org/abs/2408.03314}, 
}

@misc{wu2025inferencescalinglawsempirical,
      title={Inference Scaling Laws: An Empirical Analysis of Compute-Optimal Inference for Problem-Solving with Language Models}, 
      author={Yangzhen Wu and Zhiqing Sun and Shanda Li and Sean Welleck and Yiming Yang},
      year={2025},
      eprint={2408.00724},
      archivePrefix={arXiv},
      primaryClass={cs.AI},
      url={https://arxiv.org/abs/2408.00724}, 
}

@misc{brown2024largelanguagemonkeysscaling,
      title={Large Language Monkeys: Scaling Inference Compute with Repeated Sampling}, 
      author={Bradley Brown and Jordan Juravsky and Ryan Ehrlich and Ronald Clark and Quoc V. Le and Christopher Ré and Azalia Mirhoseini},
      year={2024},
      eprint={2407.21787},
      archivePrefix={arXiv},
      primaryClass={cs.LG},
      url={https://arxiv.org/abs/2407.21787}, 
}
@misc{liang2024improvingllmreasoningscaling,
      title={Improving LLM Reasoning through Scaling Inference Computation with Collaborative Verification}, 
      author={Zhenwen Liang and Ye Liu and Tong Niu and Xiangliang Zhang and Yingbo Zhou and Semih Yavuz},
      year={2024},
      eprint={2410.05318},
      archivePrefix={arXiv},
      primaryClass={cs.LG},
      url={https://arxiv.org/abs/2410.05318}, 
}

@misc{muennighoff2025s1simpletesttimescaling,
      title={s1: Simple test-time scaling}, 
      author={Niklas Muennighoff and Zitong Yang and Weijia Shi and Xiang Lisa Li and Li Fei-Fei and Hannaneh Hajishirzi and Luke Zettlemoyer and Percy Liang and Emmanuel Candès and Tatsunori Hashimoto},
      year={2025},
      eprint={2501.19393},
      archivePrefix={arXiv},
      primaryClass={cs.CL},
      url={https://arxiv.org/abs/2501.19393}, 
}

@inproceedings{zheng2024llamafactory,
  title={LlamaFactory: Unified Efficient Fine-Tuning of 100+ Language Models},
  author={Yaowei Zheng and Richong Zhang and Junhao Zhang and Yanhan Ye and Zheyan Luo and Zhangchi Feng and Yongqiang Ma},
  booktitle={Proceedings of the 62nd Annual Meeting of the Association for Computational Linguistics (Volume 3: System Demonstrations)},
  address={Bangkok, Thailand},
  publisher={Association for Computational Linguistics},
  year={2024},
  url={http://arxiv.org/abs/2403.13372}
}

@misc{sumers2024cognitivearchitectureslanguageagents,
      title={Cognitive Architectures for Language Agents}, 
      author={Theodore R. Sumers and Shunyu Yao and Karthik Narasimhan and Thomas L. Griffiths},
      year={2024},
      eprint={2309.02427},
      archivePrefix={arXiv},
      primaryClass={cs.AI},
      url={https://arxiv.org/abs/2309.02427}, 
}

@misc{li2024longcontextllmsstrugglelong,
      title={Long-context LLMs Struggle with Long In-context Learning}, 
      author={Tianle Li and Ge Zhang and Quy Duc Do and Xiang Yue and Wenhu Chen},
      year={2024},
      eprint={2404.02060},
      archivePrefix={arXiv},
      primaryClass={cs.CL},
      url={https://arxiv.org/abs/2404.02060}, 
}

@misc{hsieh2024middlecalibratingpositionalattention,
      title={Found in the Middle: Calibrating Positional Attention Bias Improves Long Context Utilization}, 
      author={Cheng-Yu Hsieh and Yung-Sung Chuang and Chun-Liang Li and Zifeng Wang and Long T. Le and Abhishek Kumar and James Glass and Alexander Ratner and Chen-Yu Lee and Ranjay Krishna and Tomas Pfister},
      year={2024},
      eprint={2406.16008},
      archivePrefix={arXiv},
      primaryClass={cs.CL},
      url={https://arxiv.org/abs/2406.16008}, 
}

@misc{showdown2025,
  title        = {The Showdown Computer Control Evaluation Suite},
  author       = {General Agents Team},
  year         = {2025},
  url          = {https://github.com/generalagents/showdown},
}

@misc{browserable2024webvoyager,
  author       = {{Browserable Team}},
  title        = {WebVoyager Benchmark: Browserable's State-of-the-Art Results},
  year         = {2024},
  url          = {https://www.browserable.ai/blog/web-voyager-benchmark},
  note         = {Accessed: 2025-06-03}
}

@misc{browseruse2024sota,
  author       = {{Browser Use Team}},
  title        = {SOTA Technical Report on WebVoyager Evaluation},
  year         = {2024},
  url          = {https://browser-use.com/posts/sota-technical-report},
  note         = {Accessed: 2025-06-03}
}

@misc{skyvern2024stateofart,
  author       = {{Skyvern Team}},
  title        = {Skyvern 2.0: State-of-the-Art Web Navigation with 85.8\% on WebVoyager Eval},
  year         = {2024},
  url          = {https://blog.skyvern.com/skyvern-2-0-state-of-the-art-web-navigation-with-85-8-on-webvoyager-eval},
  note         = {Accessed: 2025-06-03}
}

@misc{emergence2024agente,
  author       = {{Emergence AI}},
  title        = {Agent-E: A Text-Only Agent Outperforming Multimodal Agents},
  year         = {2024},
  url          = {https://www.emergence.ai/blog/agent-e-sota},
  note         = {Accessed: 2025-06-03}
}

@misc{hcompany2024runnerh,
  author       = {{H Company}},
  title        = {A Research Update: Runner H 0.1 Performance on WebVoyager},
  year         = {2024},
  url          = {https://www.hcompany.ai/blog/a-research-update},
  note         = {Accessed: 2025-06-03}
}

@misc{claude2024computeruse,
  author       = {{Anthropic}},
  title        = {Claude 3.5 Sonnet: Computer Use Evaluation Results},
  year         = {2024},
  url          = {https://www.anthropic.com/index/claude-computer-use},
  note         = {Accessed: 2025-06-03}
}

@misc{openai2024operator,
  author       = {{OpenAI}},
  title        = {OpenAI Operator with GPT-4o for Web Interaction Tasks},
  year         = {2024},
  url          = {https://openai.com/blog/gpt-4o},
  note         = {Accessed: 2025-06-03}
}

@inproceedings{hu2022lora,
  title     = {LoRA: Low-Rank Adaptation of Large Language Models},
  author    = {Hu, Edward J. and Shen, Yelong and Wallis, Phillip and Allen-Zhu, Zeyuan and Li, Yuanzhi and Wang, Shean and Chen, Weizhu},
  booktitle = {International Conference on Learning Representations (ICLR)},
  year      = {2022},
  url       = {https://arxiv.org/abs/2106.09685}
}

@article{shinn2023reflexion,
  title={Reflexion: Language agents with verbal reinforcement learning},
  author={Shinn, Noah and Liu, Jeffrey and Du, Yilun and Abbeel, Pieter},
  journal={arXiv preprint arXiv:2303.11366},
  year={2023}
}

@article{huang2023language,
  title={Language models as tool makers: Teaching LLMs to invent and use tools for reasoning},
  author={Huang, Yuhuai and Chen, Andy T and Bubeck, Sébastien and Zhang, Yi},
  journal={arXiv preprint arXiv:2305.17126},
  year={2023}
}

@article{pan2023automated,
  title={Automated debugging of language model programs},
  author={Pan, Zifan and Liu, Rose E and Scherlis, David and others},
  journal={arXiv preprint arXiv:2305.13266},
  year={2023}
}
}

\newpage
\appendix
\section*{Appendix}
\addcontentsline{toc}{section}{Appendix}
\subsection*{7.1 Prompting}

We share below the prompts that \websight\ uses to achieve its results. Through fine prompt engineering, we iterated to the following versions that we found to work best.

\subsubsection*{7.1.1 Planning Agent}

The planning agent is only used at the beginning of \websight's operations to determine the plan it will execute. This agent's system prompt is as follows:

\begin{quote}
\texttt{You are a web automation planner. Your job is to break down web tasks into simple steps that a browser can follow.\\\\
Key points:\\
- Break tasks into basic steps\\
- Keep steps clear and direct\\
- Account for page loading\\\\
Keep your plans simple and focused on the main goal.}
\end{quote}

\noindent The system prompt establishes context and instructions for the planner. The agent's user prompt for task specific instructions is as follows:

\begin{quote}
    \texttt{
    Create a detailed plan for a browsing agent to complete the following task. Break it down into specific, actionable steps.\\\\
    Task: \{task\}\\\\
    For each step, include:\\
    1. The specific action to take, referring to specific elements on the page\\\\
    Format your response as a numbered list of steps. Be specific about URLs, element types, and expected outcomes.
    Respond in this format:\\
    <step> STEP GOES HERE </step>\\
    <step> STEP GOES HERE </step>\\
    ...
    }
\end{quote}

\subsubsection*{7.1.2 Reasoning Agent}

The reasoning agent determines next steps and executes fine details between interactions of the browsers. This agent's system prompt is as follows:
\begin{quote}
    \texttt{You are a web automation agent using ReAct framework. Your goal: complete tasks efficiently and handle failures gracefully.\\\\
    CRITICAL RULES:\\
    - Analyze screenshot carefully before each action\\
    - Use specific selectors and exact text\\
    - Wait for dynamic content when needed\\
    - Try alternatives if primary approach fails\\
    - When you're done, return "FINISHED" and then your final response\\
    - Don't scroll unless absolutely necessary\\\\
    RESPONSE FORMAT:\\
    <reasoning>Brief analysis of current state and why this action advances the goal</reasoning>\\
    <action>Specific action (e.g., "Click the blue 'Login' button", "Type 'user@email.com' in email field", "Navigate to https://site.com")</action>\\\\
    IF YOU ARE FINISHED:\\
    <reasoning>Your reasoning here</reasoning>\\
    <action>FINISHED + your final response</action>\\\\
    Handle common patterns: loading states, forms, modals, authentication. Each Action should be a single step and be atomic (e.g. don't click on a button and then type in a text field).}
\end{quote}

\noindent The system prompt establishes a lot of the key rules and format and instructions for the reasoner. The agent's user prompt for state specific instructions is as follows:

\begin{quote}
    \texttt{TASK: \{plan\}\\
    HISTORY: \{history\}\\
    SCREENSHOT: [Current page state]\\\\
    ANALYSIS REQUIRED:\\
    1. What's on screen right now?\\
    2. What's the next logical step toward the goal?\\
    3. What could go wrong and how to handle it?\\\\
    <reasoning>\\
    Current state: [What you see]\\
    Next step: [Why this action moves toward goal]\\
    Risk mitigation: [Backup plan if this fails]\\
    </reasoning>\\\\
    <action>[Precise action instruction]</action>\\
    BE CONCISE. BE ACCURATE. HANDLE EDGE CASES.}
\end{quote}

\subsubsection*{7.1.3 Vision \websight-7B Agent}
The vision agent must deliver a simple actionable task to the browser wrapper to complete. Therefore, we ensure it is in the action space through the prompt. As this is not a chat model, there is no system prompt, and only one prompt that is displayed below: 

\begin{quote}
    \texttt{
    You are a GUI agent. You are given a task and your action history, with screenshots. You need to perform the next action to complete the task.\\\\
    \#\# Output Format\\
    ```\\
    Thought: ...\\
    Action: ...\\
    ```\\\\
    \#\# Action Space\\\\
    click(point='<point>x1 y1</point>')\\
    left\_double(point='<point>x1 y1</point>')\\
    right\_single(point='<point>x1 y1</point>')\\
    drag(start\_point='<point>x1 y1</point>', end\_point='<point>x2 y2</point>')\\
    hotkey(key='ctrl c') \# Split keys with a space and use lowercase. Also, do not use more than 3 keys in one hotkey action.\\
    type(content='xxx') \# Use escape characters \textbackslash\textbackslash', \textbackslash\textbackslash\textbackslash", and \textbackslash\textbackslash n in content part to ensure we can parse the content in normal python string format. If you want to submit your input, use \textbackslash\textbackslash n at the end of content. \\
    scroll(point='<point>x1 y1</point>', direction='down or up or right or left') \# Show more information on the `direction` side.\\
    wait() \#Sleep for 5s and take a screenshot to check for any changes.\\
    finished(content='xxx') \# Use escape characters \textbackslash\textbackslash ', \textbackslash\textbackslash ", and \textbackslash\textbackslash n in content part to ensure we can parse the content in normal python string format.\\\\
    \#\# Note\\
    - Use \{language\} in `Thought` part.\\
    - Write a small plan and finally summarize your next action (with its target element) in one sentence in `Thought` part.\\\\
    DO NOT REPEAT ACTIONS. If an action is not successful, try something else. If you've already clicked on something, don't click on it again, either try another action or do something else like typing. \\\\
    If you are stuck or a website is blocked, use the finished action to stop the agent with the argument "STUCK"\\\\
    \#\# User Instruction\\
    \{instruction\}
    }
\end{quote}

\subsection*{7.2 Fine-tuning}\label{sec:apxfinetune}
\noindent The finetuning process completes a majority of the learning very quickly. There are a multitude of reasons for this. First, websites are a large part of the UI-TARS training data, and are not too far different from mobile and desktop application data. Therefore, there is not too much learning needed. Second, UI-TARS also supports English as it's outputs are a mix of English and Chinese, so fine tuning the language for interpretability should be quick.\\

\begin{figure}[h]\
\centering
\includegraphics[width=0.48\textwidth]{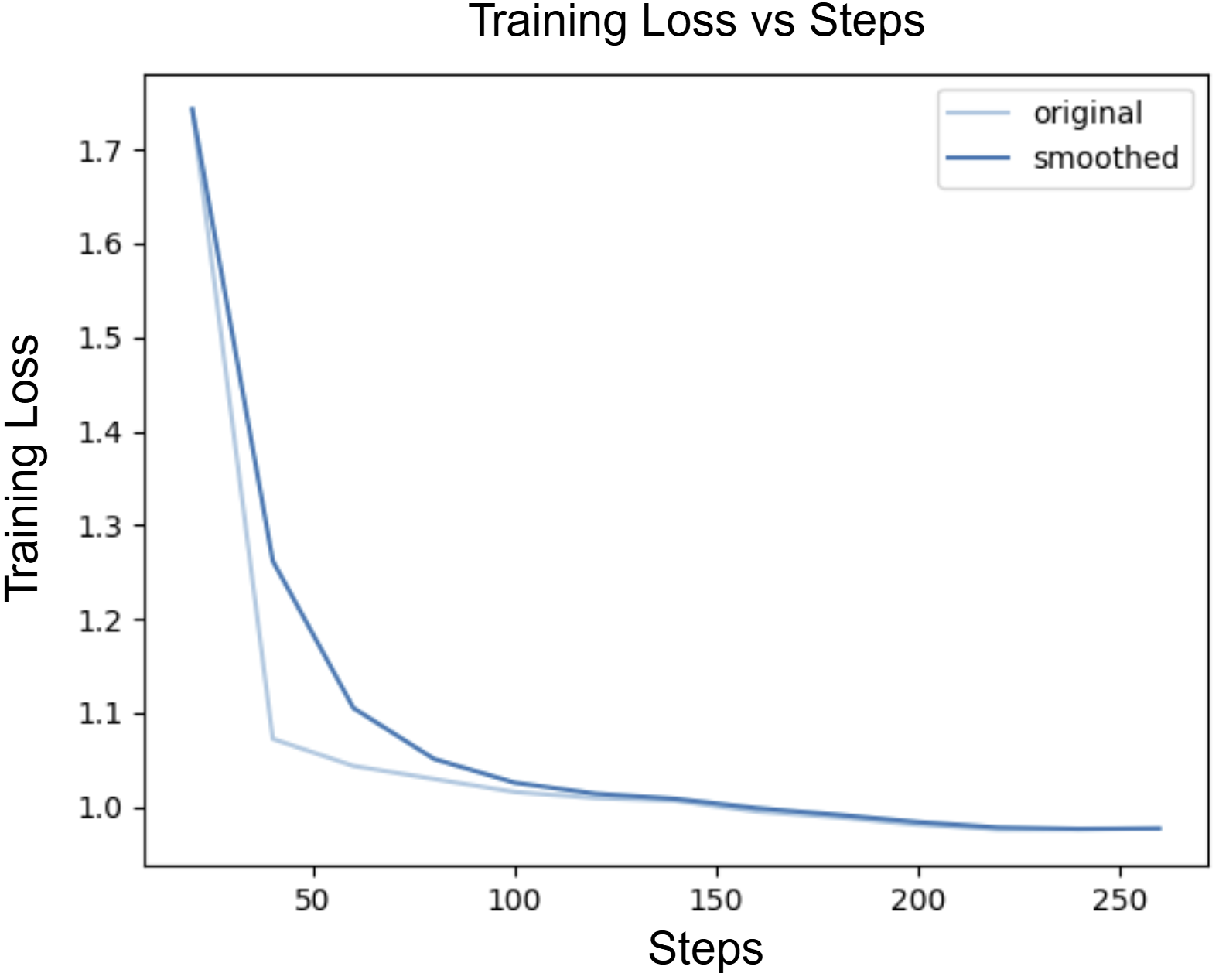}
\caption{UI-TARS LoRA Fine Tuning process Training Loss vs Steps Graph}
\label{fig:tloss}
\end{figure}

During the LoRA fine-tuning process, several special tokens were introduced to extend the model’s capabilities for structured inputs, tool usage, and multimodal interactions. These include \texttt{<|im\_start|>} (151644) and \texttt{<|im\_end|>} (151645) for denoting the start and end of instruction-style prompts, as well as \texttt{<|object\_ref\_start|>} (151646) and \texttt{<|object\_ref\_end|>} (151647) for marking object references in input sequences. Visual and spatial data are supported through tokens such as \texttt{<|box\_start|>} (151648), \texttt{<|box\_end|>} (151649), \texttt{<|quad\_start|>} (151650), and \texttt{<|quad\_end|>} (151651), along with \texttt{<|vision\_start|>} (151652), \texttt{<|vision\_end|>} (151653), and \texttt{<|vision\_pad|>} (151654) for vision input boundaries and padding.

To handle image and video input formats, the tokens \texttt{<|image\_pad|>} (151655) and \texttt{<|video\_pad|>} (151656) were added. For managing tool interactions, the tokens \texttt{<tool\_call>} (151657) and \texttt{</tool\_call>} (151658) were introduced. Code-related modifications are supported via \texttt{<|fim\_prefix|>} (151659), \texttt{<|fim\_middle|>} (151660), \texttt{<|fim\_suffix|>} (151661), and \texttt{<|fim\_pad|>} (151662), enabling more flexible handling of function-in-the-middle completions. Finally, \texttt{<|repo\_name|>} (151663) and \texttt{<|file\_sep|>} (151664) were included to represent code repository metadata and file separation, respectively. The standard \texttt{<|endoftext|>} token (151643) remains as a sentinel for sequence termination. These additions enable the fine-tuned model to operate effectively in a variety of structured, interactive, and multimodal scenarios.
\newpage

\subsection*{7.3 \textbf{\websight-7B} Failure Modes on Showdown/Clicks}\label{sec:apxfailure}

Predicted click location highlighted in \color{red}{red} and ground truth bounding box highlighted in \color{blue}{blue}

\begin{figure}[h]
    \centering
    \includegraphics[width=\linewidth]{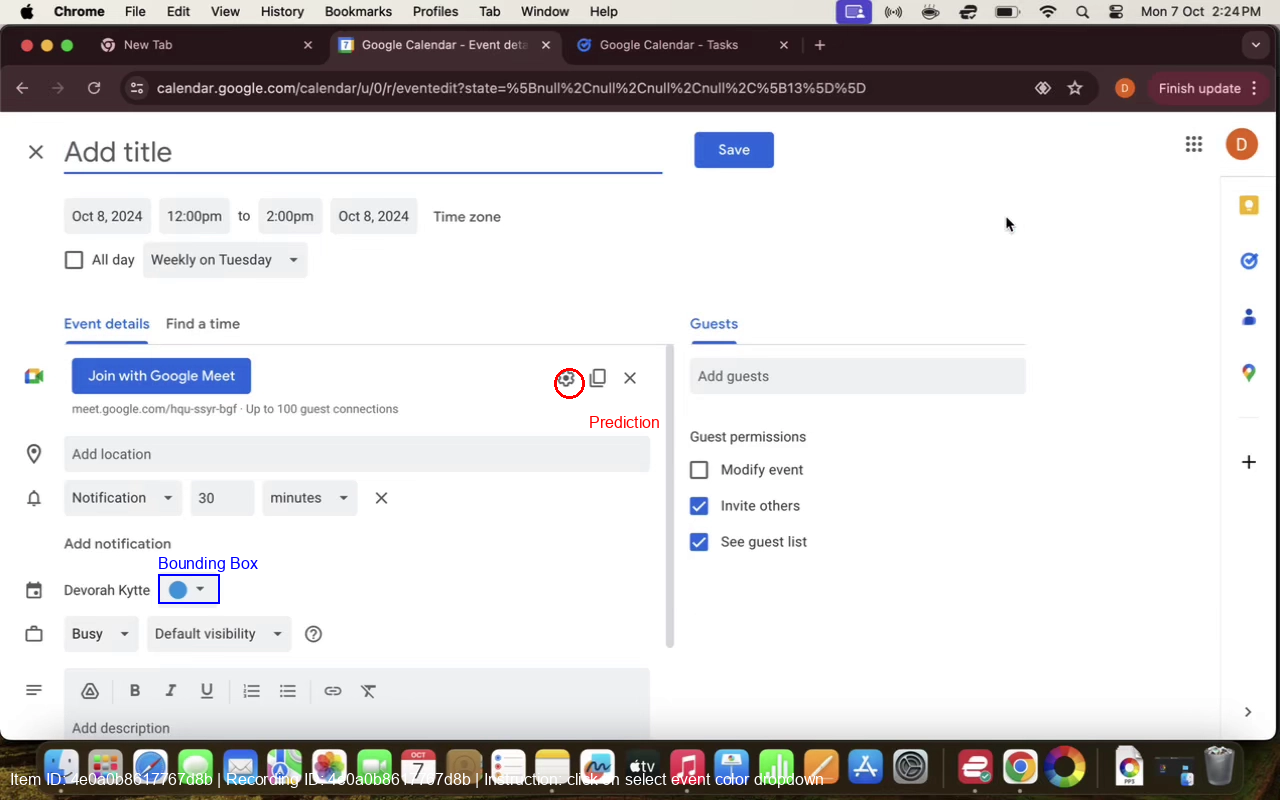}
    \caption{Click on select event color dropdown}

    \label{fig:enter-label}
\end{figure}

\begin{figure}[h]
    \centering
    \includegraphics[width=\linewidth]{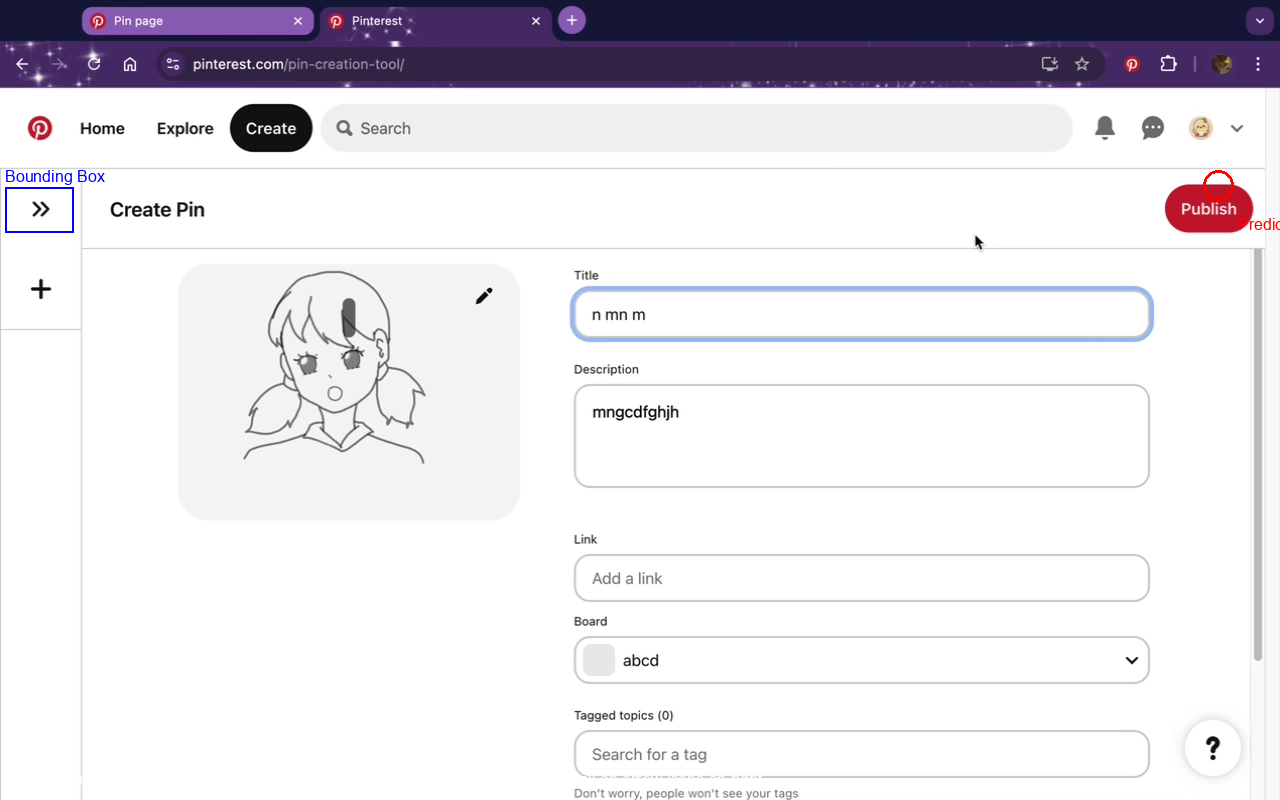}
    \caption{Click on arrow icons on right}
    \label{fig:enter-label}
\end{figure}
    
\begin{figure}[h]
    \centering
    \includegraphics[width=\linewidth]{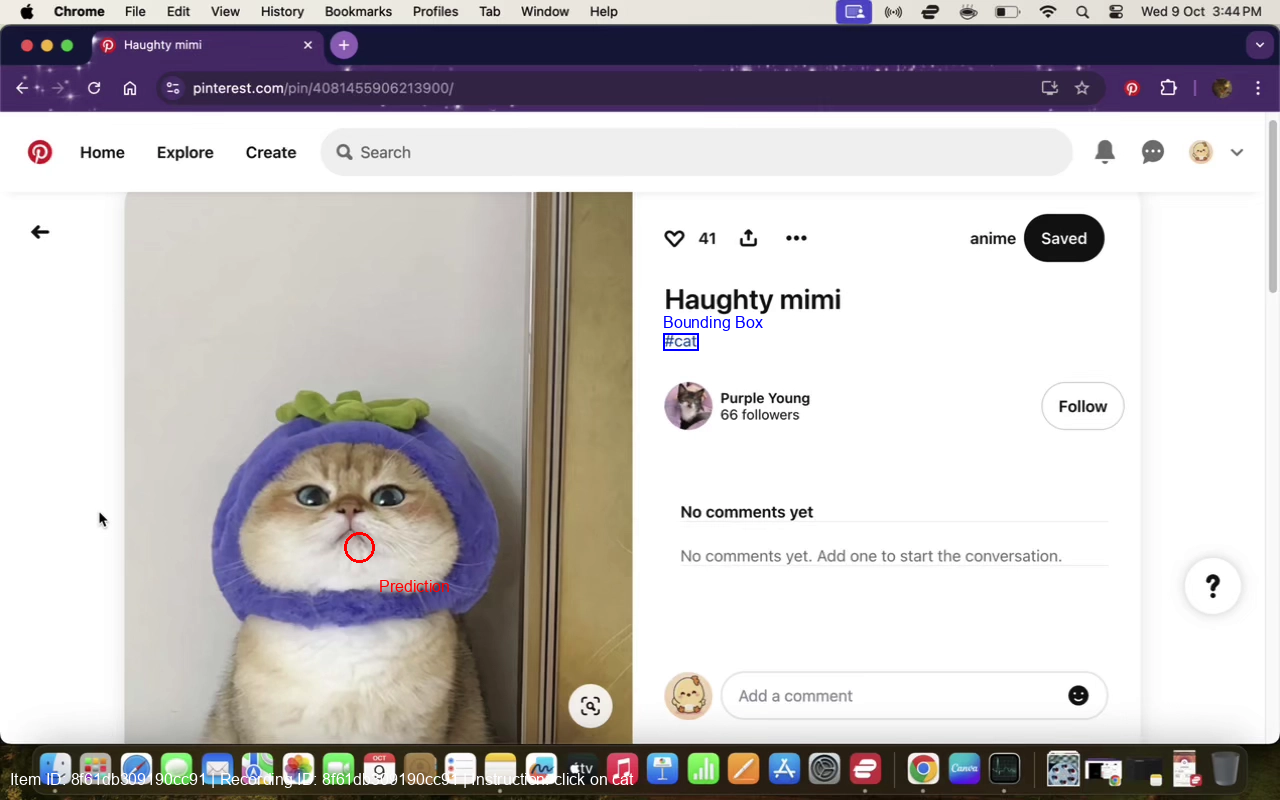}
    \caption{Click on cat}

        \label{fig:enter-label}
\end{figure}

\begin{figure}[h]
    \centering
    \includegraphics[width=\linewidth]{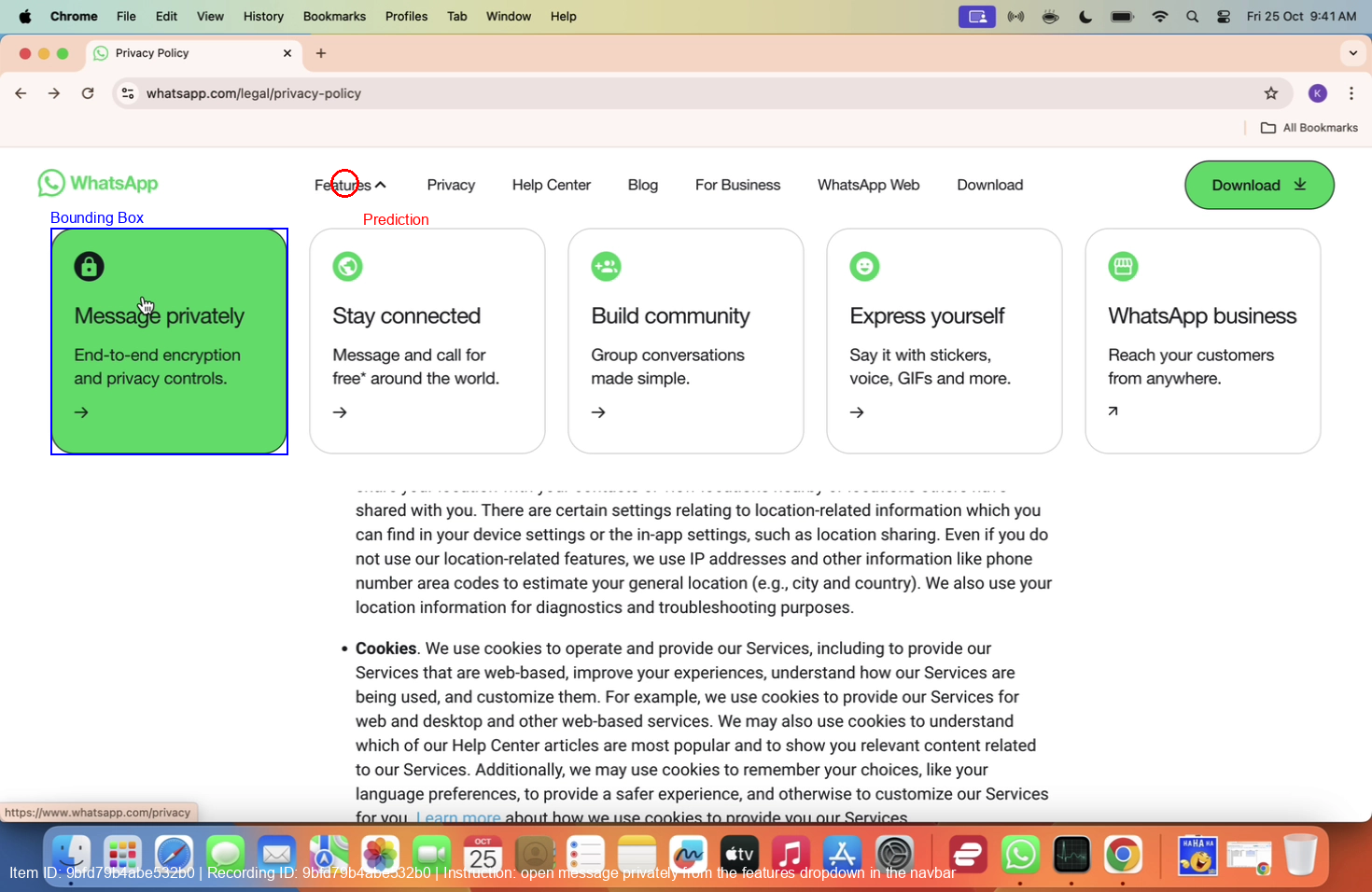}
    \caption{Open message privately from the features dropdown in the navbar}
    \label{fig:enter-label}
\end{figure}

\begin{figure}[h]
    \centering
    \includegraphics[width=\linewidth]{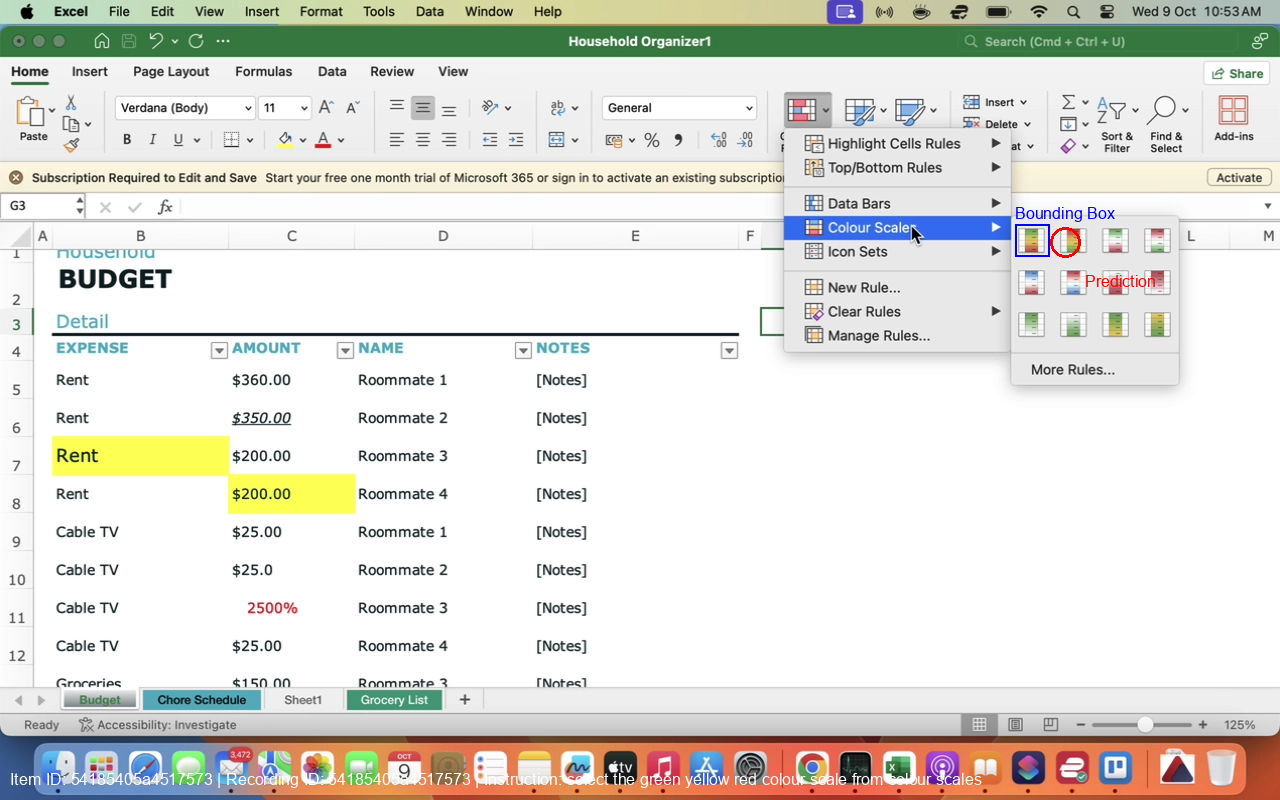}
    \caption{Select the green yellow red colour scale from colour scales}
    \label{fig:enter-label}
\end{figure}

\begin{figure}[h]
    \centering
    \includegraphics[width=\linewidth]{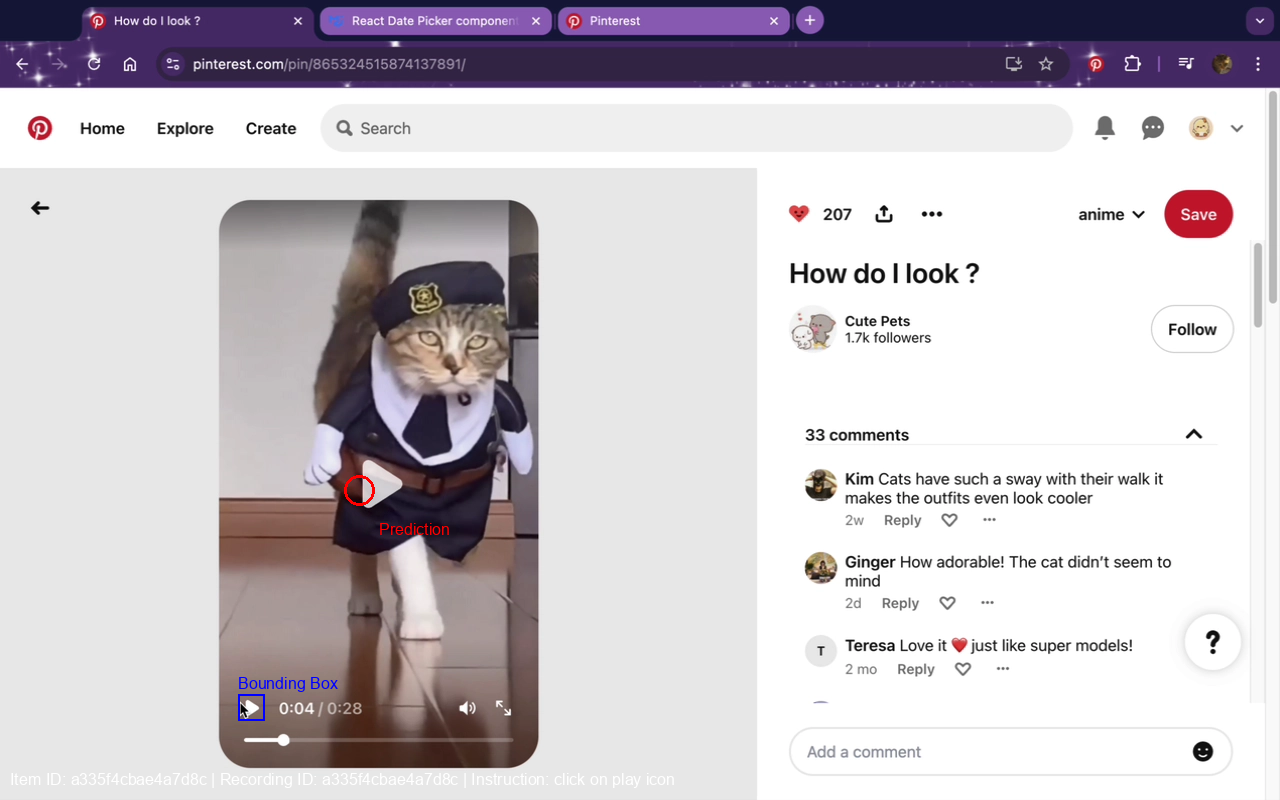}
    \caption{Click on play icon}
    \label{fig:enter-label}
\end{figure}
\end{document}